\documentclass{article}

\usepackage[preprint]{neurips_2024}

\usepackage[utf8]{inputenc} 
\usepackage[T1]{fontenc}    
\usepackage{hyperref}       
\usepackage{url}            
\usepackage{booktabs}       
\usepackage{amsfonts}       
\usepackage{nicefrac}       
\usepackage{microtype}      
\usepackage{xcolor}         
\usepackage{graphicx}
\usepackage{wrapfig}
\usepackage{listings}
\usepackage{caption}
\usepackage{subcaption}
\usepackage{ulem}
\usepackage{amsmath}
\usepackage{bbding}

\usepackage[capitalize]{cleveref} 
\usepackage[linesnumbered,ruled,vlined]{algorithm2e}
\graphicspath{{Figs/}}

\title{UniGS: Modeling Unitary 3D Gaussians for Novel View Synthesis from Sparse-view Images}

\author{%
Jiamin Wu$^{1,2}$\thanks{This work is done during an internship in the International Digital Economy Academy (IDEA).}~~, Kenkun Liu$^{3}$\thanks{These authors contributed equally to this work.}~~, Yukai Shi$^{2,4}$, Xiaoke Jiang$^2$\thanks{Corresponding author}~~, Yuan YAO$^1$~~, Lei Zhang$^2$\\
$^1$Hong Kong University of Science and Technology \\
$^2$International Digital Economy Academy (IDEA)\\
$^3$The Chinese University of Hong Kong, Shenzhen \\
$^4$Tsinghua University
} 

\begin{document}

\maketitle

\begin{abstract}
In this work, we introduce \textbf{UniGS}, a novel 3D Gaussian reconstruction and novel view synthesis model that predicts a high-fidelity representation of 3D Gaussians from arbitrary number of posed sparse-view images.
Previous methods often regress 3D Gaussians locally on a per-pixel basis for each view and then transfer them to world space and merge them through point concatenation.
In contrast, Our approach involves modeling unitary 3D Gaussians in world space and updating them layer by layer.
To leverage information from multi-view inputs for updating the unitary 3D Gaussians, we develop a DETR (DEtection TRansformer)-like framework, which treats 3D Gaussians as queries and updates their parameters by performing multi-view cross-attention (\textbf{MVDFA}) across multiple input images, which are treated as keys and values.
This approach effectively avoids `ghosting' issue and allocates more 3D Gaussians to complex regions.
Moreover, since the number of 3D Gaussians used as decoder queries is independent of the number of input views, our method allows arbitrary number of multi-view images as input without causing memory explosion or requiring retraining.
Extensive experiments validate the advantages of our approach, showcasing superior performance over existing methods quantitatively (improving PSNR by 4.2 dB when trained on Objaverse and tested on the GSO benchmark) and qualitatively. The code will be released at \href{https://github.com/jwubz123/UNIG}{\color{blue}{https://github.com/jwubz123/UNIG}}.
\end{abstract}

\section{Introduction}
3D object reconstruction and novel view synthesis (NVS) play a crucial role in the fields of computer vision and graphics. The construction of detailed 3D structures from 2D images has applications including robotics, augmented reality, virtual reality, medical imaging, and beyond. Recently, 3D Gaussian Splatting (3D GS) \citep{3d-gs}, as a semi-explicit representation, has demonstrated great efficiency and high-quality rendering performance for NVS.

\begin{figure*}
    \centering
    \includegraphics[width=1.0\textwidth]{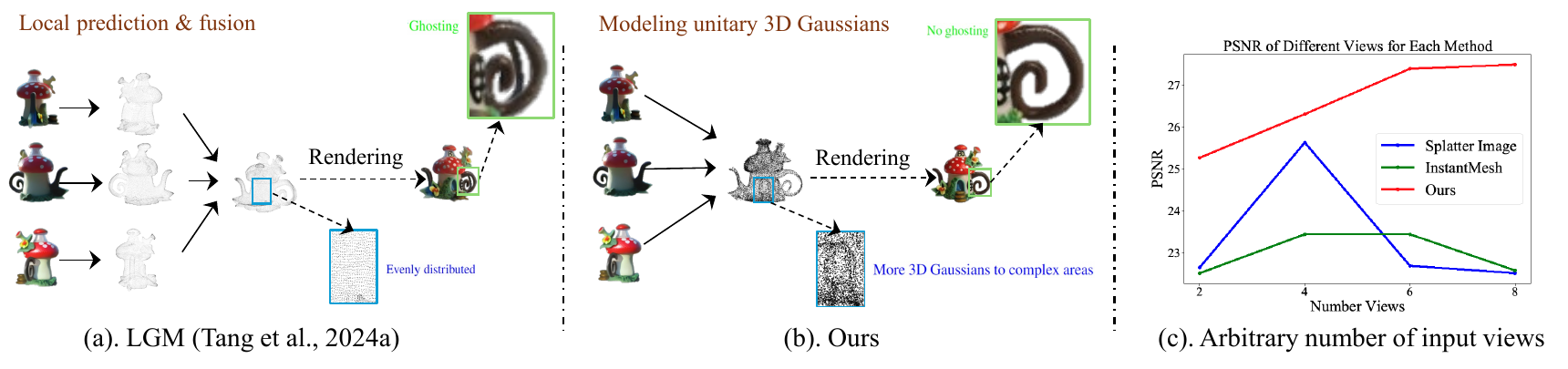}
    
    \caption{(a) Previous methods like LGM \citep{LGM} initially predict 3D Gaussians for each pixel for each view and then merge them to get the final 3D Gaussians, resulting in a \textcolor{green}{`ghosting'} issue. Moreover, the 3D Gaussians are \textcolor{blue}{evenly distributed} for both simple and complex regions, while there should be more 3D Gaussians for the complex regions. (b) In contrast, our approach utilizes a unitiary set of 3D Gaussians, projecting them onto each view and gathering information across views through a global optimization strategy. Our model effectively avoids the \textcolor{green}{`ghosting'} problem and assigns \textcolor{blue}{more 3D Gaussians to complex areas} (such as the `door' in the image). (c) Our approach supports an arbitrary number of inputs views without requiring retraining and the performance does not deduced. (Trained on 4 input views and tested on 2 to 8 views).}

    \label{fig: teaser}
\end{figure*}

Despite the advantages of 3D GS, they often perform inadequately when faced with sparse-view inputs (no more than 10 views). Some previous works \citep{SplatterImage,LGM,mvgamba, gslrm, grm} propose to exploit large-scale multi-view image datasets and train feed-forward models that learn to reconstruct 3D Gaussians of objects from sparse-view images in a single forward process.
However, many recent feed-forward methods based on 3D GS employ a `local prediction and fusion' approach in which they predict 3D Gaussians for every pixel locally in each input view under the corresponding camera coordinates. Subsequently, the predicted 3D Gaussians are transformed into the same world coordinate system by the input view camera poses and are concatenated to get the final 3D Gaussian representation. This strategy results in \textcolor{blue}{even distribution} of the positions of 3D Gaussians in both simple and complex regions  (\cref{fig: teaser}) because the local predication on the even distributed pixels on each view. Moreover, the fusion of the 3D Gaussians from each view may encounter the challenge of `\textcolor{green}{ghosting}', i.e. the predicted per-pixel 3D Gaussians from different views cannot align to others after concatenation. As depicted in \cref{fig: teaser}(a), an extra part of the rendered object can be obviously observed. This issue arises from that the estimated positions of 3D Gaussians in each view may be imprecise, so that the merging of 3D Gaussians cannot align well. Another limitation of previous methods is that they require the same number of input views during both training and inference stages; otherwise, their performance would notably deteriorate (\cref{fig: teaser}(c)), constraining practical applications where the quantity of input views is variable.


To address these issues, we propose a \textbf{Uni}tary 3D \textbf{G}aussians (UniGS) representation for sparse-view reconstruction. Inspired by Deformable DETR (DEtection TRansformer) \citep{DETR} that treats the position and properties of bounding box (Bbox) as queries, we develop a DETR-like framework that treats the unitary 3D Gaussians as queries and updates them layer by layer with multi-view image features as keys and values in cross-attention. 

To efficiently leverage multi-view image features, we introduce multi-view deformable cross-attention (\textbf{MVDFA}). Global queries are first defined for the unitary 3D Gaussians within world space. These global queries contribute to generating queries, keys and values for the deformable cross-attention. On the one hand, the global queries undergo modulation by camera parameters for each view to derive view-specific queries, considering that the input contains multiple views but the 3D Gaussians are defined unitarily. Specifically, inspired by \citep{stylegan, openlrm}, the global queries are subject to a linear transformation using weights and biases derived from a multilayer perceptron (MLP) conducted on the camera parameters of each view to obtain the view-specific queries. One the other hand, the global queries are processed through an MLP to predict 3D Gaussians. These 3D Gaussians (centers of them) are projected onto each input view to generate projected points, serving as reference points in deformable DETR within each view. Subsequently, the view-specific queries are updated by deformable cross-attention with keys and values being image features sampled surrounding the reference points for each view. Finally, global queries are updated by all view-specific queries merged through a weighted summation, where the weights are determined by an MLP operating on the view-specific queries. 

With the above design, utilizing unitary 3D Gaussians that avoid both per-pixel local optimization and direct fusion of 3D Gaussians from each view, our model effectively addresses the `ghosting' problem, allocates more 3D Gaussians to complex regions, and supports arbitrary number of input views in inference without retraining.

In summary, our contributions are as follows:
\begin{itemize}
    \item We propose UniGS, a novel 3D object reconstruction and NVS algorithm which introduces a unitary set of 3D Gaussians in world space, enabling all input views to contribute to a unified 3D representation.
    \item We present MVDFA to efficiently utilize multi-view image features by conducting cross-attention within each view and fuse them into the same set of 3D Gaussians.
    \item Both quantitative and qualitative experiments are conducted for evaluation. Our proposed method achieves the state-of-the-art performance on the commonly-used object benchmark GSO.
\end{itemize}

\paragraph{3D reconstruction from images}
Recently, various methods have been explored to reconstruct detailed 3D object from limited viewpoints. 
\citep{LGM, one2345, openlrm, dreamgaussian, ddim} view the problem as an image-conditioned generation task. Leveraging pretrained generative models like \cite{stablediffusion}, they achieve realistic renderings of novel views. However, maintaining view consistency in the generated images is challenging and diffusion models require longer time to generate a single image with denoising process, thus limiting their applicability in real-time scenarios. Moreover, the fusion of information from multiple images remains a non-trivial challenge. Neural Radiance Field (NeRF) \citep{nerf} has gained prominence as a widely used 3D representation \citep{pixelnerf, FWD, FE-NVS, visionnerf}. Techniques such as InstantMesh \citep{instantmesh}, which combine triplane and NeRF for 3D reconstruction, have demonstrated promising results. 
\citep{pixelnerf, IBRNet, MVSNeRF} extract pixel-aligned feature embeddings from multiple views and merge them using MLPs. 
However, due to its slow rendering speed, NeRF is being supplanted by a new, super-fast, semi-implicit representation—3D Gaussian Splatting (3D GS) \citep{3d-gs}. SplatterImage \citep{SplatterImage}, LGM \citep{LGM} based on 3D Gaussian Splatting, typically handle each input view independently and naively concatenate the resulting 3D Gaussian assets from each view. This method suffers from a lack of information exchange among different views, resulting in inefficient utilization of 3D Gaussians and being view inconsistency. Furthermore, these methods are unable to accommodate an arbitrary number of views as input.

\paragraph{Deformable Transformer in 3D} \label{sec: 3d dfa}
DFA3D \citep{dfa3d} and BEVFormer \citep{bevformer} are introduced to address the feature-lifting challenge in 3D detection and autonomous driving tasks. They achieve notable performance enhancements by employing a deformable Transformer to bridge the gap between 2D and 3D. DFA3D initially uses estimated depth to convert 2D feature maps to 3D, sampling around reference points for deformable attention in each view. However, the 3D sampling point design causes all projected 2D points to represent a singular point, neglecting view variations. BEVFormer \citep{bevformer} regards the Bird's-Eye-View (BEV) features as queries, projecting the feature onto each input view. The Spatial Cross-Attention facilitates the fusion of BEV and image spaces, though challenges persist sampling 4 height values per pillar in the BEV feature for selecting 3D reference points may limit coverage, posing challenges in accurate keypoint selection for the model. When contrasting DFA3D and BEVFormer with our MVDFA, a commonality lies in projecting onto 3D regression targets to extract data from various image perspectives. However, our model diverges by employing camera modulation to differentiate queries across views, enabling more specific information retrieval. 
\section{Related Work} \label{sec: related_works}
\begin{figure*}
    \centering
    \includegraphics[width=0.85\textwidth]{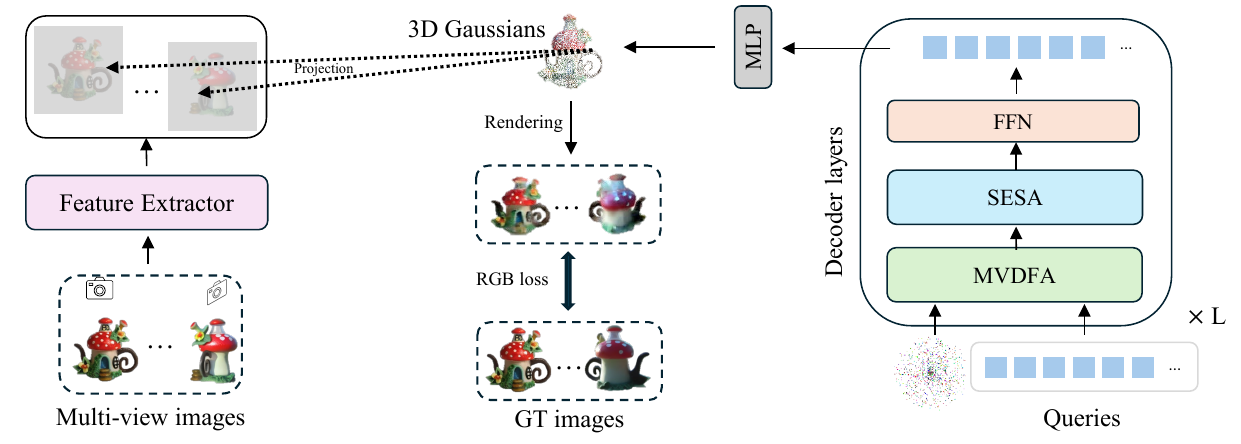}
    
    \caption{UniGS: Queries are updated by $L$ decoder layers with multi-view image features extracted by the feature extractor. 3D Gaussians are regressed from the queries by an MLP in each layer. Subsequently, they are passed into the next layer and projected onto each view to derive reference points.  MVDFA: multi-view deformable attention in \cref{sec: decoder}. SESA: spatial efficient self-attention in \cref{sec: space_efficient_self_attn}. The dashed arrow means that the centers of 3D Gaussians are projected to multi-view feature maps to retrieve the most related features.}
    \label{fig: overview}
    
\end{figure*}
\paragraph{3D reconstruction from images}
Recently, various methods have been explored to reconstruct detailed 3D object from limited viewpoints. 
\citep{One-2-3-45++, one2345, dreamgaussian, ddim} view the problem as an image-conditioned generation task. Leveraging pretrained generative models like \cite{stablediffusion}, they achieve realistic renderings of novel views. However, diffusion models require longer time to generate 3D with multi-step denoising process, thus limiting their applicability in real-time scenarios. 
Recent methodologies that rely on a single forward process for 3D reconstruction, utilizing Neural Radiance Field (NeRF) \citep{nerf} as a robust 3D representation, have demonstrated effective performance in the field of 3D reconstruction. \citep{pixelnerf, FWD, FE-NVS, visionnerf, instant3d, instantnerf, meshformer, meshlrm, TripoSR, instantmesh, pixelnerf, IBRNet, MVSNeRF, liu2024comprehensive, xiong2024mvhumannet}.
However, due to the slow rendering speed of NeRF, it is being supplanted by a new, super-fast, semi-implicit representation—3D Gaussian Splatting (3D GS) \citep{3d-gs}. Triplane-Gaussian \citep{triplane-gs}, Gamba \citep{gamba}, and LeanGaussian \citep{wu2024dig3d} make promising results on single image 3D reconstruction. When it comes to the parts that does not appear in the input image, the models can not constuct them well.
Various techniques such as SplatterImage \citep{SplatterImage}, LGM \citep{LGM}, pixelSplat \citep{pixelsplat}, MVSplat \citep{mvsplat}, GS-LRM \citep{gslrm}, and GRM \citep{grm} have extended the application of 3D Gaussian Splatting to multi-view scenarios. 
In these approaches, each input view is processed to estimate 3D Gaussians specific to the view, followed by a simple concatenation of the resulting 3D Gaussian assets from all views, resulting in the `ghosting' problem and evenly distribute 3D Gaussians on the object. Such design demands substantial computational resources, particularly as the number of views grows, the number of Gaussians scales linearly with the number of views. Furthermore, these methods are unable to accommodate an arbitrary number of views as input.  As an concurrent work, GeoLRM \citep{geolrm} also exploit the power of transformer and cross-attention, but they still use the discrete voxel representation which may cause high computational cost and potentially get lower accuracy.

\paragraph{Deformable Transformer in 3D} 
Deformable DETR \citep{DETR} and its following works \citep{dab-detr, dfa3d, detr-matching, taptr, detr-matching, taptr, wu2024dig3d, geolrm, dino} makes successful attempts in many fields.
DFA3D \citep{dfa3d} and BEVFormer \citep{bevformer} are introduced to address the feature-lifting challenge in 3D detection and autonomous driving tasks. They achieve notable performance enhancements by employing a deformable Transformer to bridge the gap between 2D and 3D. 
DFA3D initially uses estimated depth to convert 2D feature maps to 3D, sampling around reference points for deformable attention in each view. However, the 3D sampling point design causes all projected 2D points to represent a singular point, neglecting view variations. BEVFormer \citep{bevformer} regards the Bird's-Eye-View (BEV) features as queries, projecting the feature onto each input view. The Spatial Cross-Attention facilitates the fusion of BEV and image spaces, though challenges persist sampling 4 height values per pillar in the BEV feature for selecting 3D reference points may limit coverage, posing challenges in accurate keypoint selection for the model. When contrasting DFA3D and BEVFormer with our MVDFA, a commonality lies in projecting onto 3D regression targets to extract data from various image perspectives. However, our model employ camera modulation to differentiate queries across views, enabling more specific information retrieval. 


\section{Methods}
\begin{figure*}
\centering
\begin{subfigure}[b]{0.4\textwidth}
    \includegraphics[width=\textwidth]{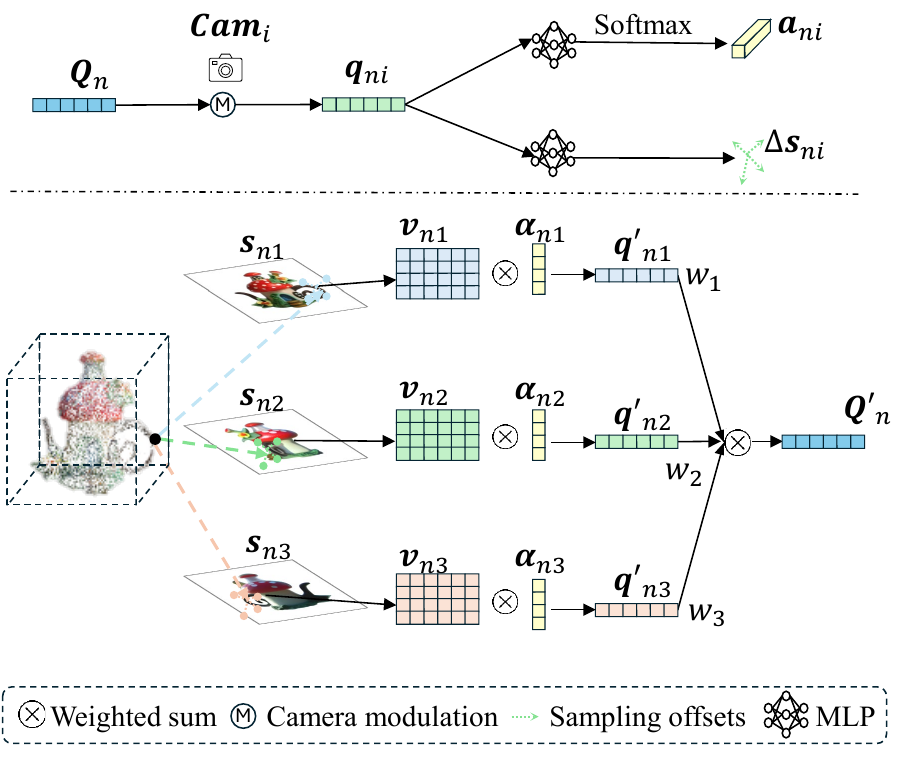}
    \caption{MVDFA on the $n$-th 3D Gaussian}
    
\end{subfigure}
\hfill 
\begin{subfigure}[b]{0.5\textwidth}
\begin{lstlisting}[mathescape=true, basicstyle=\ttfamily\tiny]
def MVDFA($\mathbf{F}$, $K$, $\pi$, $\mathbf{Q}$, $\boldsymbol{\mu}$):
    # Prepare camera embedding
    camera=concat($K$, $\pi$).flatten() #[B, I, 16]
    cam_embed=MLP(camera) #[B, I, C]
    # Modulate query by cameras, [B, I, N, C]
    shift, scale=MLP(cam_embed).chunk(2)  
    $\mathbf{q}$=LayerNorm($\mathbf{Q}$)*(1+scale)+shift #[B, I, N, C]
    $\boldsymbol{\alpha}$=softmax(Linear($\mathbf{q}$)) #Attn score [B, I, N, Ns]
    # Sampling points
    $\Delta \mathbf{s}$=Linear($\mathbf{q}$) #[B, I, N, Ns, 2]
    $\mathbf{P}$=pinhole_proj(camera, $\boldsymbol{\mu}$) #[B, I, N, 1, 2]
    $\mathbf{s}$=$\mathbf{P}$+$\Delta \mathbf{s}$ #[B, I, N, Ns, 2]
    # Weighted sum of view-specific queries
    $\mathbf{V}$=Linear($\mathbf{F}$) # [B, I, H*W, C]
    $\mathbf{v}$=grid_sample($\mathbf{V}$, $\mathbf{s}$)  #[B, I, Ns, C]
    $\mathbf{q'}$=($\boldsymbol{\alpha} \cdot \mathbf{v}$).sum(-1) #[B, I, N, C]
    $w$=sigmoid(Linear($\mathbf{q'}$)) #[B, I, C]
    $\mathbf{Q'}$=($w \cdot \mathbf{q'}$).sum(-2) #[B, N, C]
    return $\mathbf{Q^{'}}$ #[B, N, C]

\end{lstlisting}
\caption{Pseudo code}
\end{subfigure}

\caption{MVDFA: $\mathbf{Q}_n$ denotes the $n$-th unitary queries while $\textbf{q}_{ni}$ denotes the $n$-th query on the $i$-th view modulated by the $i$-th camera $\mathbf{Cam}_i$. Sampling offsets $\Delta\mathbf{s}_{ni}$ and attention score $\boldsymbol{\alpha}_{ni}$ derived by conducting linear transformation on  $\textbf{q}_{ni}$. The sampling offsets are utilized to sample image features at the sampling points $\mathbf{s}_{ni} = \mathbf{P}_{ni} + \Delta\mathbf{s}_{ni}$, where  $\mathbf{P}_{ni}$ is the reference point derived by projecting the $n$-th 3D Gaussian. After that, $\mathbf{s}_{ni}$ is utilized to sample image features serving as values $\mathbf{v}_{ni}$. These values are employed to update the view-specific queries by attention scores $\boldsymbol{\alpha}_{ni}$. The unitary queries are refined by the weighted sum of updated view-specific queries $\textbf{q}'_{ni}$, where $w_i$ is the weight calculated by a linear layer on $\textbf{q}'_{ni}$. $B$ is batch size, $I$ is the number of views, $C$ is the hidden dimension, $N$ is the number of Gaussians, $\textit{pinhole\_proj}$ is the projection from 3D to 2D with the pinhole model. $\mathbf{F}$ is the image feature with height $H$ and width $W$. $\textbf{K}$ and $\boldsymbol{\pi}$ are camera intrinsics and extrinsics, respectively.} 
\label{fig: mvdfa}

\end{figure*}
Before introducing details for UniGS, we first give the preliminaries for 3D GS and deformable cross-attention. (see \cref{sec: prelim}).
After that, as illustrated in \cref{fig: overview}, our model follows an encoder-decoder framework. All input images undergo processing through an image encoder and a cross-view attention module to extract multi-view image features. For the decoder, we employ unitary 3D Gaussian representation, which define a unitary set of 3D Gaussians in the world space no matter how many input views are given.
Each 3D Gaussian is then projected onto each view to query relevant features and update their respective parameters by query refinement decoder with multi-view deformable attention (MVDFA) (see \cref{sec: decoder}). Spatially efficient self-attention is utilized to reduce computational and memory costs, enabling the utilization of more 3D Gaussians for object reconstruction (see \cref{sec: space_efficient_self_attn}). The training objective is finally introduced. (see \cref{sec: loss}).
\subsection{Prelimenaries} \label{sec: prelim}
\paragraph{3D GS} \label{sec: 3d-gs}
3D GS \citep{3d-gs} is a novel rendering method that can be viewed as an extension of point-based rendering methods \citep{3d-gs, 3d_gs_survey}. They use 3D Gaussians as effective 3D representation for efficient differentiable rendering.
The $N$ 3D Gaussians representing an object can be described by $\mathbf{G} = \{\boldsymbol{\mathrm{SH}}, \boldsymbol{\mu}, \boldsymbol{\sigma}, \mathbf{R}, \mathbf{S}\}$.
The color of 3D Gaussians is represented by spherical harmonics SH $\in \mathbb{R}^{N \times 12}$ while the geometry is described by the center positions $\boldsymbol{\mu} \in \mathbb{R}^{N \times 3}$, shapes (rotation $\mathbf{R} \in \mathbb{R}^{N \times 3 \times 3}$ and scales $\mathbf{S} \in \mathbb{R}^{N \times 3}$ \footnote{The shape of 3D Gaussian ellipsoids can be described by the covariance matrix $\boldsymbol{\Sigma}$, which can be optimized through a combination of rotation and scaling for each ellipsoid as $\boldsymbol{\Sigma} = \mathbf{R}\mathbf{S}\mathbf{S}^T\mathbf{R}^T$.}), and opacity $\boldsymbol{\sigma} \in \mathbb{R}^{N \times 1}$ of ellipsoids \citep{EWASplatting, 3d-gs}. In our model, the goal for 3D Gaussian reconstruction is to estimate these parameters for a object consists of N 3D Gaussians by a feed-forward network.

\paragraph{Deformable cross-attention} \label{sec: deformable cross_attn}
Deformable cross-attention, a novel mechanism introduced in Deformable DETR  \citep{DETR}, enhances traditional cross-attention by dynamically adjusting spatial sampling locations based on input features. It is initially designed for detection tasks (bounding box center and size prediction). It defines bbox centers as reference points $\mathbf{P}$ and its content features as queries $\mathbf{Q}$. Keys and values in deformable attention are image features sampled surrounding the reference points at the sampling points $\mathbf{s}$. To get the sampling points, learnable offsets $\Delta\mathbf{s}$ are predicted by MLP on $\mathbf{Q}$, and are added to reference points to get the sampling points $\mathbf{s}$. These learned offsets focus on modeling the most important regions of input feature maps to refine queries. Then the sampling points are utilized to sample image features by grid sampling with bilinear interpolation to get the keys and values of the cross-attention, enabling precise feature attention. The queries are then refined by the image features through deformable attention. Deformable DETR provides a way capturing detailed, context-aware relationships within input data, enhancing overall performance.

\subsection{Overview} \label{sec: our model}

\paragraph{Feature extractor encoder} \label{sec: encoder}
To extract image features $\mathbf{F}$ from multi-view input, we utilize UNet \citep{unet, songunet}, a widely employed feature extractor in 3D reconstruction tasks, as demonstrated in \cite{LGM, SplatterImage}. To enhance the network's understanding of the complete 3D object, multi-view cross-attention is employed to transfer information among views right after the UNet block, activated when the number of input views exceeds one. In this context, each input view acts as queries, while the concatenation of the remaining views serves as keys and values.
To efficiently enable cross-attention across all views, we employ shifted-window attention, as introduced in the Swin Transformer \citep{swinT}. This mechanism reduces interactions by focusing on tokens within a local window, effectively reducing memory usage for large input sequences. 

\paragraph{Query refinement decoder} \label{sec: decoder}
In the decoder module, we define a fixed number of queries $\mathbf{Q} \in \mathbb{R}^{N \times C}$ with $N$ and $C$ denoting the number of Gaussians and the hidden dimension. The queries are utilized to model 3D Gaussians $\boldsymbol{G}$ including the center $\boldsymbol{\mu}$, opacity $\boldsymbol{\sigma}$, rotation $\mathbf{R}$, scaling $\mathbf{S}$, and Spherical Harmonics $\mathbf{SH}$. Note that queries and 3D Gaussians are one-to-one corresponded.
As depicted in \cref{fig: overview}, the queries go through multiple decoder layers, each including a multi-view deformable attention (MVDFA) (\cref{sec: mvdfa}) mechanism to leverage image features, a spatial efficient self-attention (SESA) (\cref{sec: space_efficient_self_attn}) layer for inter-Gaussian interactions, and a feed-forward network (FFN). The functionality of a decoder layer can be summarized by \cref{eq: decoder_layer}, where $\mathbf{F}$ represents image features from different views,  $\mathbf{Q}^l$ is the queries in th $l$-th layer, and $\mathbf{P}^l$ is reference points in the $l$-th layer.
\begin{equation} \label{eq: decoder_layer}
    \mathbf{Q}^{l+1} = \mathrm{FFN}(\mathrm{SESA}(\mathrm{MVDFA}(\mathbf{Q}^{l}, \mathbf{P}^l, \mathbf{F}))
\end{equation}
Finally, queries are processed through an MLP to compute $\Delta \mathbf{G} = \mathrm{MLP}(\mathbf{Q})$ for updating the 3D Gaussian parameters: $\mathbf{G}' = \mathbf{G} + \Delta \mathbf{G}$ \footnote{For the updating of 3D Gaussian parameters, rotation is updated by multiplication, while other parameters are updated by addition.}.
The initialization for the positions of the center of 3D Gaussians, which make sure that they are not too distant from the ground truth or outside the field of view to gurantee the training convergence, are given in \cref{app: analysis}.

\subsection{Multi-view deformable attention (MVDFA)} \label{sec: mvdfa}
\paragraph{View-specific queries generation}
As depicted \cref{fig: mvdfa} (a), in 3D, the set of queries associated with 3D Gaussian paremeters $\mathbf{G}$ are defined unitarily, while cross-attention is defined in multiple image planes. Therefore, teh unitary queries should be adjusted to suit each view individually. To solve this issue, we employ camera modulation with the adaptive layer norm (adaLN) \citep{openlrm, stylegan, stylegan1, stylegan2} to generate view-specific queries. More specifically, as shown in \cref{fig: mvdfa}(a), the unitary queries $\textbf{Q}$ are transformed linearly to get the view-specific queries ($\textbf{q}_i$ for the $i$-th view) by weights and bias deriving from conducting MLP on camera parameters. 

\paragraph{Updating view-specific queries}
Following deformable DETR (see \cref{sec: prelim}), the image feature, which are most related to the queries should be sampled surrounding the reference points. These reference points are defined as the position of queries, where in our setting, defined as the projected points in each view for center of 3D Gaussians. More specifically, given center $\boldsymbol{\mu}$ of 3D Gaussians along with the corresponding camera poses $\boldsymbol{\pi}_i$ and intrinsic parameters $\textbf{K}_i$ for the $i$-th view, we can compute UV coordinates $\mathbf{P}_i$ by projecting the center coordinates of each 3D Gaussian onto the image plane of the $i$-th input image using the pinhole camera model (see \cref{eq: pinhole}) \citep{pinhole1, pinhole2},
\begin{equation} \label{eq: pinhole}
    \mathbf{P}_i = \textbf{K}_i\boldsymbol{\pi}_i\boldsymbol{\mu}
\end{equation}.
In this context, both matrices $K_i$ and $\pi_i$ are expressed in homogeneous form. These projected points $\mathbf{P}_i$ are then regarded as the reference points for 2D deformable attention. After getting the reference points in each view, as shown in \cref{fig: mvdfa}, a linear layer is employed to predict the sampling offsets $\Delta\mathbf{s}_i$ to get the sampling points $\mathbf{s}_i$ surrounding the reference points by \cref{eq: sample_points}.
\begin{equation} \label{eq: sample_points}
    \mathbf{s}_i = \mathbf{P}_i + \Delta\mathbf{s}_i
\end{equation}
Then, we apply the grid sampling algorithm with bilinear interpolation to extract image features at these sampling points, which act as the values $\mathbf{v}$ for cross attention.
Subsequently, another linear layer is employed to predicts the attention scores $\boldsymbol{\alpha}$ of the image features $\mathbf{v}$ at the sampling points $\mathbf{s}$. Finally, for each input view, we compute the updated queries by the dot product between the attention scores $\boldsymbol{\alpha}$ and the sampled values $\mathbf{v}$.

\paragraph{Fusion of view-specific queries}
After getting the refined queries for each view $\textbf{q}'$, the unitary queries $\textbf{Q}'$ are then computed as a weighted sum of individual view queries, with the weights calculated using an linear layer on the view-specific queries (see \cref{eq: weight_sum}).
\begin{equation} \label{eq: weight_sum}
    \textbf{Q}' = \sum_i w_i \textbf{q}_i', w_i = \mathrm{MLP}(\textbf{q}_i')
\end{equation}
Detailed pseudo code for MVDFA is available in \cref{fig: mvdfa}(b).

\paragraph{Insights}
With MVDFA, all views contribute to unitary 3D Gaussians, emphasizing the most relevant features. This strategy effectively alleviates the view inconsistency issue and is computationally more efficient.

\subsection{Spatial efficient self-attention (SESA)} \label{sec: space_efficient_self_attn}
Apart from MVDFA, self-attention in each decoder layer is also important to get the information across all 3D Gaussians, However, when the number of 3D Gaussians $N$ is large, the self-attention is computationally expensive. To tackle this problem, inspired by \cite{pvt}, we introduce SESA that reduce the number of keys and values while keeping the number of queries unchanged during self-attention. Insights behind the design is that updating each Gaussian with information from all others may not always be essential, as neighboring Gaussians often contain similar information. This selective updating strategy enables each query to be updated with a subset of related queries instead of all other queries, effectively enhancing the information exchange efficiency. To ensure enough information flow in the downsampled queries, we leverage the Fast Point Sampling (FPS) algorithm from point cloud methodologies \citep{pointnet, pointnet++}. Specifically, we employ FPS Gaussian centers $\boldsymbol{\mu}$ to identify the most distant points from all 3D Gaussians' centers and use the corresponding queries as keys and values in the self-attention. With such strategy, our model optimize memory usage while guaranteeing essential information sharing among Gaussians. Additional details are in \cref{app: space_efficient_self_attn}.

\subsection{Training objective} \label{sec: loss}
Building upon prior 3D Gaussian-based reconstruction approaches, we leverage the differentiable rendering implementation by \cite{3d-gs} to generate RGB images from the 3D Gaussians produced by our model. For each object, we render 4 input views and 8 additional views (12 views in total) for supervision. Furthermore, aligning with the methodologies (\citep{openlrm, LGM}), we employ a RGB loss in \cref{eq: loss}, which consists of both a mean square error loss $\mathcal{L}_{\text{MSE}}$ and a VGG-based LPIPS (Learned Perceptual Image Patch Similarity) loss \citep{LIPIS} $\mathcal{L}_{\text{LPIPS}}$ to guide the rendered views. Here $I_{pd}$ represents the rendered views supervised by the ground truth images $I_{gt}$. $\lambda$ is the scalar weiths on the LPIPS loss.
\begin{equation} \label{eq: loss}
    \mathcal{L} = \mathcal{L}_{\text{MSE}}(I_{pd}, I_{gt}) + \lambda\mathcal{L}_{\text{LPIPS}}(I_{pd}, I_{gt})
\end{equation}




\section{Experiments} \label{sec: exp}
\paragraph{Dataset} \label{sec: dataset}
We utilize a refined subset of the Objaverse LVIS dataset \citep{objaverse} for training and validation. 
The training dataset included input rendering images captured from fixed viewpoints (front, back, left, right) and supervised by 32 random views spanning elevations between -30 to 30 degrees. The resolution of the rendered images was downscaled to $128 \times 128$.
To evaluate our model, we conducted tests on the Google Scanned Objects (GSO) benchmark with fixed-view inputs (e.g., front, left, back, right) at 0 degrees elevation, tested on 32 random views with elevations ranging from 0 to 30 degrees. More details can be found in \cref{app: datasets}.

\paragraph{Evaluation metric} \label{sec: metric}
We compute the peak signal-to-noise ratio (PSNR), structural similarity index (SSIM) \citep{ssim}, and perceptual distance (LPIPS) \citep{lpips} between the rendered images and the ground truth to evaluate the NVS quality. 
Additionally, we offer visual representations for both the rendered images and the 3D Gaussian centers as point clouds. 

\paragraph{Implementation details are shown} in \cref{sec: exp_setting}.

\subsection{Comparison with state-of-the-art methods} \label{sec: quan_res}

\begin{figure}[!t]
    \centering
    \includegraphics[width=0.99\textwidth]{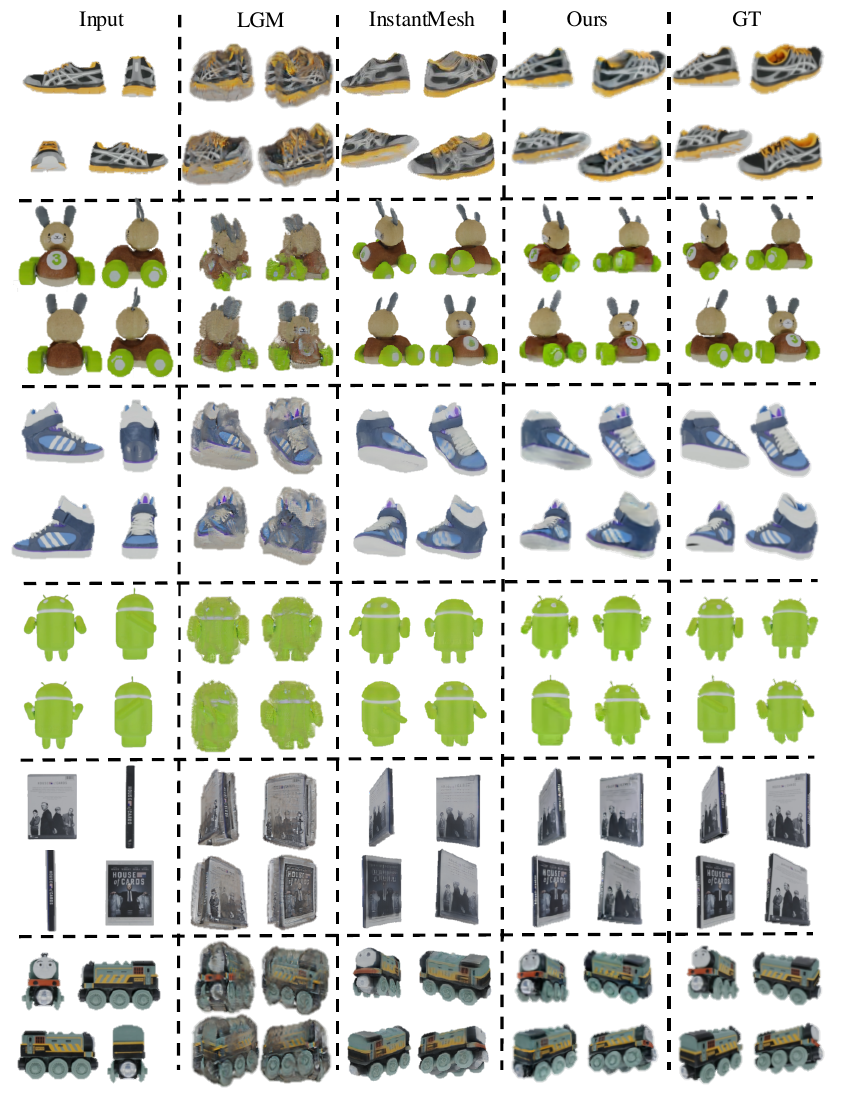}
    
    \caption{Novel views on GSO dataset for inputting 4 views with resolution 128.}
    \label{fig: fixed_vis_128}
    
\end{figure}

\begin{table}
\small
  \caption{Quantitative results for inputting 4 views on GSO dataset. *The results of MV-Gamba and GS-LRM are cited from the paper. `NA' means not reported in the paper. Resolution is 128.}
  
  \label{tab: 4_fix_view_results}
  \centering
  \begin{tabular}{llll}
    \toprule
       Method & PSNR $\uparrow$     & SSIM $\uparrow$ & LPIPS $\downarrow$ \\
    \midrule
    Splatter Image \citep{SplatterImage} & 25.6241 & 0.9151 & 0.1517  \\
    LGM (Small) \citep{LGM} & 17.4810 & 0.7829 & 0.2180   \\
    LGM (Large) \citep{LGM}  & 26.2487 & 0.9249 & 0.0541 \\
    InstantMesh \citep{instantmesh} & 23.0177 & 0.8893 & 0.0886 \\
    GeoLRM* \citep{geolrm} & 22.8400 & 0.8510 & NA \\
    MV-Gamba* \citep{mvgamba}  &  26.2500  & 0.8810 & 0.0690  \\
    GRM (Res-512)* \citep{grm} & 30.0500 & 0.9060 & 0.0520 \\
    GS-LRM (Res-256)* \citep{gslrm}  &  29.5900 & 0.9440 & 0.0510  \\
    \midrule
    Our Model & \textbf{30.4245} & \textbf{0.9614} & \textbf{0.0422}  \\
    \bottomrule
  \end{tabular}
  
\end{table}
\paragraph{Quantity results}
We evaluate recent multi-view reconstruction models using 4 views as input. In \cref{tab: 4_fix_view_results}, LGM and InstantMesh were evaluated using the provided checkpoints, with ``Small" indicating models tailored to 128 resolution with the small model and ``Large" to 256 resolution with the large model. Splatter Image \citep{SplatterImage} is retrained on the same dataset of ours as they do not provide checkpoints with 4 input views. For other methods, we cite the results reported in their paper. \cref{tab: 4_fix_view_results} showcases the performance of these methods in novel view synthesis using 4 fixed views (front, back, right, left) on the GSO dataset. Our model surpassed previous approaches in PSNR, SSIM, and LPIPS for novel view synthesis, with a significant improvement of approximately 4.2 dB in PSNR. Additional results for 6 and 8 view inputs are available in \cref{app: other_num_views}.

\begin{figure}[t]
    \centering
    \includegraphics[width=0.99\textwidth]{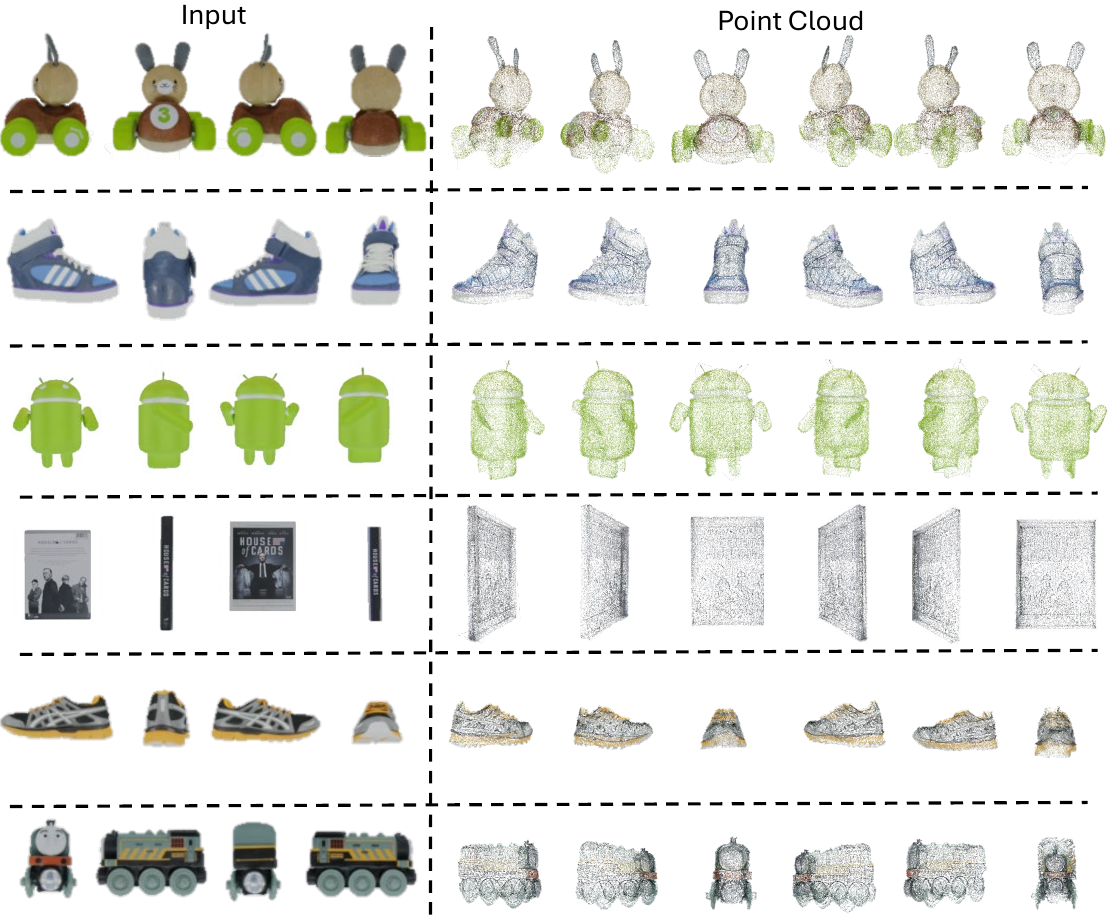}
    
    \caption{3D Gaussian center as point cloud on GSO dataset for inputting 4 views.}
    \label{fig: point_cloud}
    
\end{figure}

\paragraph{Quality results}
We present visualization results for novel view synthesis with resolution 128 in \cref{fig: fixed_vis_128}. Note that for LGM, we visualize the results generated by `small' model provided in their github. We can observe the problem of `ghosting' in LGM and the problem of lacking details in InstantMesh (see \cref{app: ghosting} for more visualization).
Further visualizations with resolution of 256 are accessible in \cref{app: more_vis}. 
Furthermore, to demonstrate that our model can handle input views with varying elevations, we also present the results for input views with random camera poses in \cref{app: random_view}. The results of comparison to scene-based models \citep{mvsplat, pixelsplat} can be found in \cref{app: mvsplat}. Comparison to the results that removing background points in previous methods are shown in \cref{app: mask}. Results obtained from inputting a single view only can be found in \cref{app: single_view}.
We also show the point cloud visualization in \cref{fig: point_cloud} underscores our model's ability to capture geometry effectively, not just rendering quality.
More visualization for the `ghosting' problem by visualize center of Gaussians from each view in different colors, are in \cref{app: more_vis} \cref{fig: view_point_cloud}. Furthermore, removing the background use masks for Splatter Image and LGM may slightly improve the performance (\cref{fig: mask}, \cref{tab: mask_results}), but still worse than our methods.

\paragraph{Inference on arbitrary number of views}
Training cost for 3D methods is large, often requiring hundreds of GPUs training multiple days. Additionally, memory costs for previous methods increase linearly with the number of views increasing, presenting challenges for training models with varying input views. Therefore, a model supporting inference with arbitary number of inputs while being trained on a fixed number of views, such as 4 views, would provide significant advantages.
Our model retains unitary 3D Gaussians in world coordinates, treating views as complementary sources without compromising overall 3D integrity. This enables adaptability to variable view counts during inference, despite training on a fixed number of views. \cref{fig: teaser} (c) showcases the results of training the model with 4 random views and testing it with different number of views. More views results are in \cref{app: other_num_views} \cref{fig: more_view}.
While other methods demonstrate satisfactory performance with 4 views during inference, their effectiveness diminishes as the view count different from 4. In contrast, our model gets increasing performance as the number of views increases. It is important that some methods can not handle variations in the number of views between the training and testing, and thus we ignore them in the figure. For the application with single input view, we show the results in \cref{fig: single_gso} and \cref{tab: single_view_result} in \cref{app: single_view}. 

\paragraph{Inference time and memory cost} \label{sec: speed}
\begin{table}
  \caption{Inference time comparison. 3D: forward time, render: rendering time, inference: time of one forward and 32 rendering. Unit in seconds.}
  
  \label{tab: speed}
  \centering

  \begin{tabular}{llll}
    \toprule
       Method & 3D $\downarrow$     & Render $\downarrow$ & Inference $\downarrow$ \\
    \midrule
    DreamGaussian & 118.3245 & 0.0038 & 118.4461 \\
    InstantMesh & \textbf{0.6049} & 0.6206 & 20.4641\\ 
    LGM & 1.6263 & 0.0090 & 1.9143 \\
    \midrule
    Our Model & 0.6939 & \textbf{0.0019} & \textbf{0.7538}  \\
    \bottomrule
  \end{tabular}
  
\end{table}
We performed inference time tests across different models, including a diffusion-based method (DreamGaussian \citep{dreamgaussian}), NeRF-based model (InstantMesh \citep{instantmesh}), previous Gaussian-based model (LGM \citep{LGM}), and our model, as shown in \cref{tab: speed}.
In contrast to previous methods that compute 3D Gaussians per pixel per input view, our model retains a single 3D Gaussian irrespective of the number of views. While conventional methods exhibit linear memory expansion with additional views or higher image resolutions, our approach sustains a consistent memory overhead or experiences slight increments due to the marginally higher cost of the image feature extractor. This design theoretically enables our model to accommodate more input views and higher resolutions for enhanced outcomes, potentially circumventing the out-of-memory limitations encountered by other methods.

\subsection{Ablation studies} \label{sec: ablation}
\begin{figure*}
    \centering
    \includegraphics[width=1.0\textwidth]{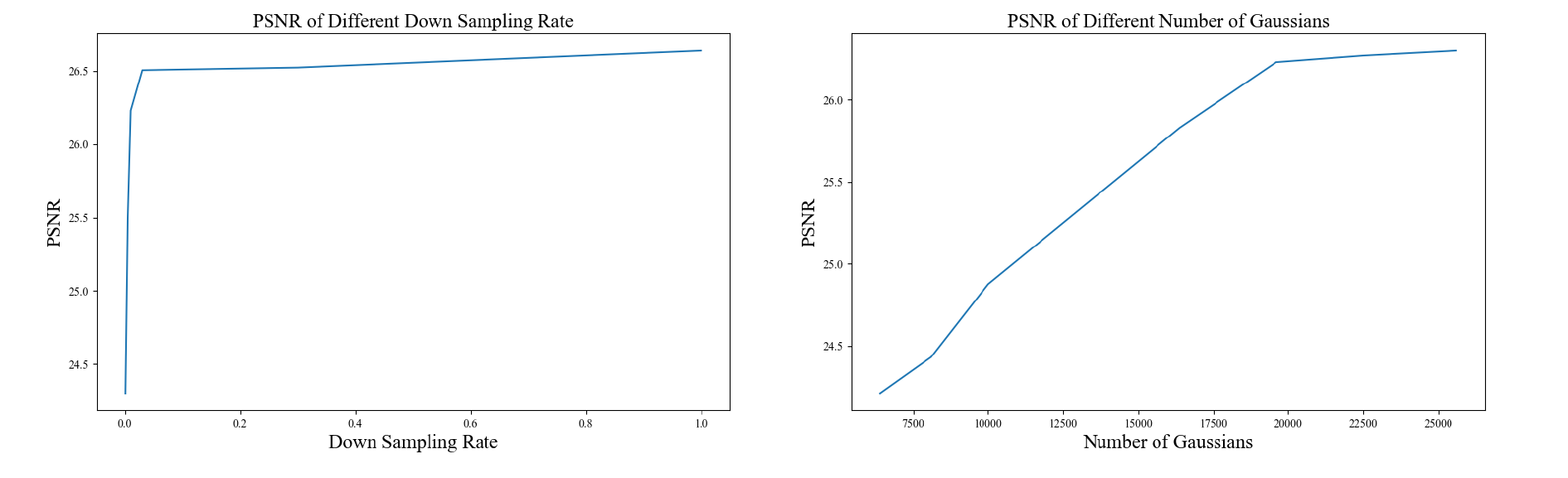}
    
    \caption{Left: PSNR with different down sampling rate in SESA. Right: PSNR with different number of Gaussians.}
    \label{fig: hyper_ablation}
    
\end{figure*}
\cref{tab: ablation_study} illustrates an ablation study that evaluates different components of the model architecture. All the experiments are evaluated on the Objaverse validation dataset. 

\paragraph{Initialization of 3D Gaussians}
As shown in \cref{tab: ablation_study}, initialize 3D Gaussians randomly (without any constraint) results in low performance. This problem arises from utilizing image features around the projected 3D Gaussian center within each image view. When projections extend beyond the image plane, the reference points may be out of the image plane and sampled images features with only value 0, which steep gradients. Initialize 3D Gaussians randomly within the cone of vision (CoV) make the performance better. In this sense, better initialization may give better final performance and therefore we initialize the 3D Gaussians by regressing them first from image features for each pixel. Details for such initialization are provided in \cref{app: analysis}. 

\paragraph{No cross-view attention in feature extractor}
Removing cross-view attention leads to a moderate decrease in performance compared to the full model. 

\paragraph{No decoder}
We do ablation study of removing the decoder, i.e., use the 3D Gaussians regressed from each pixel of the image features extracted from the feature extractor and then concatenate them as the same process of previous methods \citep{LGM, SplatterImage, gslrm}. 
As shown in \cref{app: analysis}, such design underperforms the full model. 

\paragraph{No camera modulation in MVDFA}
Furthermore, removing the camera modulation on queries or use the shared 3D sampling points instead of sampling on each view adversely impacts the results, underscoring the critical significance of this view-specific design. The full model achieves the best performance, indicating that each component contributes positively to the overall model effectiveness.

\paragraph{Ablation study on SESA}
As introduced in \cref{sec: space_efficient_self_attn}, the memory bottleneck of our model lies in the pointwise self-attention mechanism. To address this, we implement a spatially efficient self-attention technique to alleviate memory consumption. Illustrated in \cref{fig: hyper_ablation} (left), as we augment the downsampling rate of the key and value in the self-attention mechanism, the memory overhead diminishes linearly, while the PSNR reduction is not so rapid. Consequently, we opt for a downsampling rate located at the inflection point, which we determine to be 0.01, balancing memory efficiency with reconstruction quality. 

\paragraph{Number of Gaussians}
In our model, we first define a fixed number of 3D Gaussians. To show influence of the selection on the number of 3D Gaussians, we conduct ablation study on different number of 3D Gaussians. As shown in \cref{fig: hyper_ablation} (right), when the number of 3D Gaussians exceeds 19600, the increment from adding the number of Gaussians becomes flatten. Therefore, we select the number of 3D Gaussians being 19600.
Additionally, we offer details on hyperparameter selections in \cref{app: ablation_study}.

\begin{table}
\small
  \caption{Ablation study on model design.}
  
  \label{tab: ablation_study}
  \centering

  \begin{tabular}{llll}
    \toprule
    Method & PSNR $\uparrow$     & SSIM $\uparrow$ & LPIPS $\downarrow$ \\
    \midrule
    ran. init. & 12.1213 & 0.6531 & 0.6224 \\
    ran. init. in CoV & 22.6740 & 0.8711 & 0.2383 \\
    w/o cross view attention & 25.3923 & 0.9013 & 0.1007  \\
    UNet only & 25.6033 & 0.9107 & 0.0930 \\
    w/o camera modulation & 26.1328 & 0.9201 & 0.0883 \\
    3D sampling points & 25.8392 & 0.9117 & 0.0945 \\
    \midrule
    Full model  & \textbf{26.5334} & \textbf{0.9344} & \textbf{0.0667}  \\
    \bottomrule
  \end{tabular}
  
\end{table}

\paragraph{Number of input views}
We add the ablation study on the number of input views in \cref{app: ablation_study} \cref{tab: ablation_n_view}. Our model not only support 4 views as input, but also surpass previous methods on other number of input views. More ablation studies are in \cref{app: ablation}.

\subsection{Applications in 3D generation} \label{sec: gen_application}
\begin{figure}[htp]
    \centering
    \includegraphics[width=0.98\textwidth]{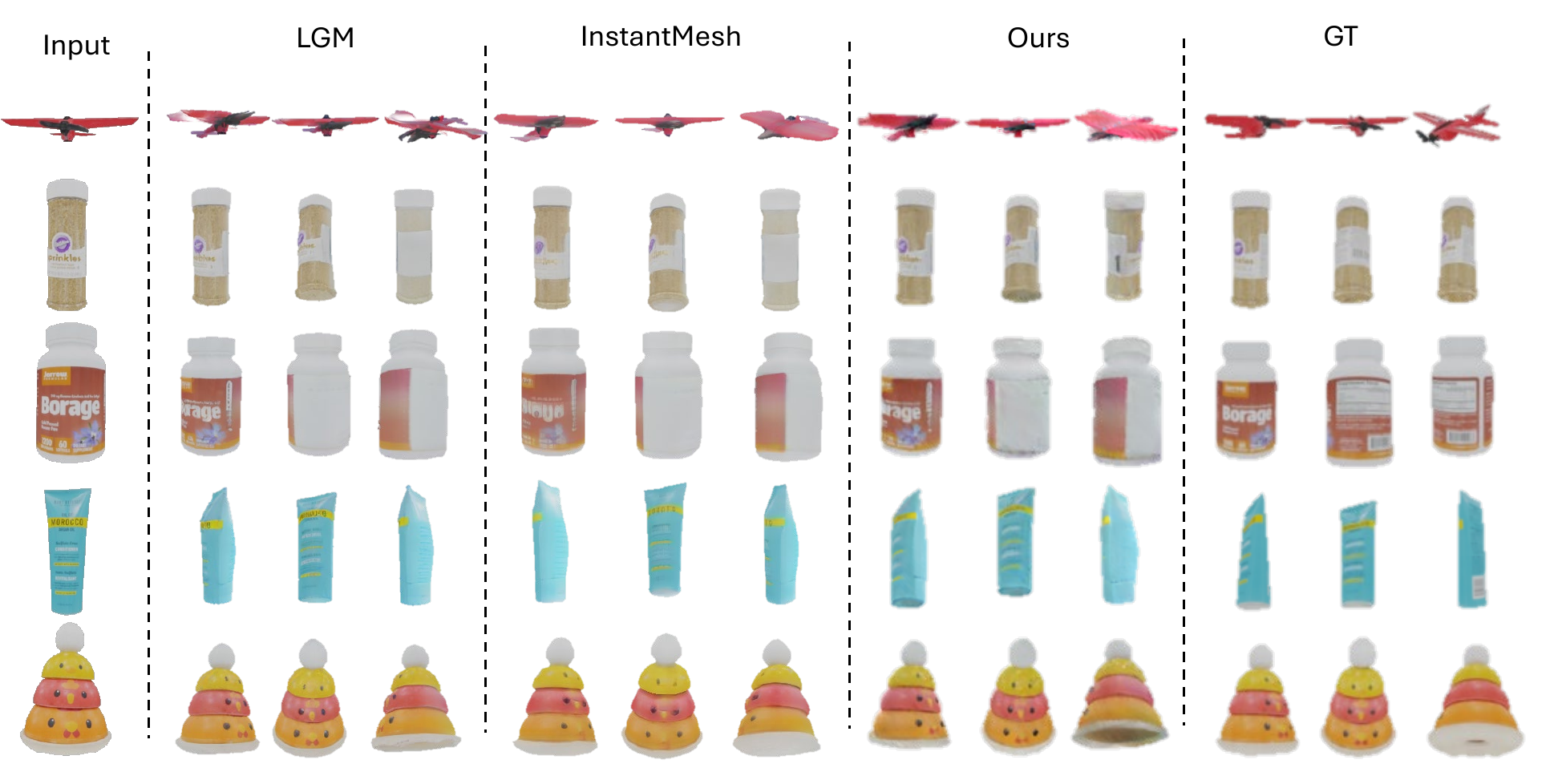}
    
    \caption{Quality for rendered novel views on GSO dataset for inputting 1 view and using ImageFusion to generate 4 views.}
    \label{fig: single_gso}
    
\end{figure}

\textbf{Image-to-3D} conversion represents a fundamental application in 3D generation. Following the methodology of LGM and InstantMesh \citep{LGM, instantmesh}, we initially leverage a multi-view diffusion model, ImageDream \citep{imagedream}, to generate four predetermined views. Subsequently, our model is utilized for 3D Gaussian reconstruction. A comparative analysis with LGM and InstantMesh is detailed in \cref{app: more_vis}.
We also showcase the quality results of our model on the GSO datase in \cref{fig: single_gso}. Additionally, we also provide the results of in-the-wild input images in \cref{app: more_vis} \cref{fig: in_the_wild}.

\paragraph{Text-to-3D} We utilize MVDream \citep{mvdream} to generate a single image from a text prompt. Subsequently, a diffusion model is employed to produce multi-view images, which are then processed by our model to derive a 3D representation. A qualitative comparison of the text-to-3D generation is presented in \cref{app: more_vis}.

\section{Conclusion and Limitation} \label{sec: conclusion}
In this paper, we introduce a novel sparse view 3D Gaussian reconstruction and NVS method. First, fixed number of unitary 3D Gaussians defined in world space together with the correponding queries are initialized, and each 3D Gaussian (center of them) is projected onto input image features extracted by a feature extractor. Then, MVDFA block is designed to do cross-attention on each view. The unitary queries are modulated to each view by the camera parameters to get the view-specific queries. The 3D Gaussians (center of them) are projected onto multi-view image plane to get the reference points. After that, image features surrounding these reference points are sampled to update the view-specific queries by deformable cross-attention in each view. Finally, the unitary queries are refined by weighted sum on the view-specific queries. Moreover, we develop a spatially efficient self-attention mechanism to minimize computational costs. With the above design, our model successfully tackling `ghosting' problem and giving more meaningful distribution of 3D Gaussians by giving more 3D Gaussians for complex areas. Moreover, our model can accept an arbitrary number of views as input without demage the performance. The experiments on GSO dataset shows the effectiveness of our model. 

\paragraph{Limitations}
Our model requires user-provided camera parameters, which are important for projection, presenting potential challenges in 3D reconstruction.
\clearpage

\bibliography{main}

\begin{thebibliography}{64}
\providecommand{\natexlab}[1]{#1}
\providecommand{\url}[1]{\texttt{#1}}
\expandafter\ifx\csname urlstyle\endcsname\relax
  \providecommand{\doi}[1]{doi: #1}\else
  \providecommand{\doi}{doi: \begingroup \urlstyle{rm}\Url}\fi

\bibitem[Cao et~al.(2022)Cao, Rockwell, and Johnson]{FWD}
Ang Cao, Chris Rockwell, and Justin Johnson.
\newblock {FWD: Real-time novel view synthesis with forward warping and depth}.
\newblock In \emph{Proceedings of the IEEE/CVF Conference on Computer Vision and Pattern Recognition (CVPR)}, pp.\  15713--15724, 2022.

\bibitem[Charatan et~al.(2024)Charatan, Li, Tagliasacchi, and Sitzmann]{pixelsplat}
David Charatan, Sizhe~Lester Li, Andrea Tagliasacchi, and Vincent Sitzmann.
\newblock pixelsplat: 3d gaussian splats from image pairs for scalable generalizable 3d reconstruction.
\newblock In \emph{Proceedings of the IEEE/CVF Conference on Computer Vision and Pattern Recognition}, pp.\  19457--19467, 2024.

\bibitem[Chen et~al.(2021)Chen, Xu, Zhao, Zhang, Xiang, Yu, and Su]{MVSNeRF}
Anpei Chen, Zexiang Xu, Fuqiang Zhao, Xiaoshuai Zhang, Fanbo Xiang, Jingyi Yu, and Hao Su.
\newblock {MVSNeRF: Fast Generalizable Radiance Field Reconstruction from Multi-View Stereo}.
\newblock In \emph{2021 IEEE/CVF International Conference on Computer Vision (ICCV)}, Oct 2021.
\newblock \doi{10.1109/iccv48922.2021.01386}.
\newblock URL \url{http://dx.doi.org/10.1109/iccv48922.2021.01386}.

\bibitem[Chen \& Wang(2024)Chen and Wang]{3d_gs_survey}
Guikun Chen and Wenguan Wang.
\newblock {A Survey on 3D Gaussian Splatting}.
\newblock \emph{ArXiv}, 2024.

\bibitem[{Chen, Yuedong and Xu, Haofei and Zheng, Chuanxia and Zhuang, Bohan and Pollefeys, Marc and Geiger, Andreas and Cham, Tat-Jen and Cai, Jianfei}(2024)]{mvsplat}
{Chen, Yuedong and Xu, Haofei and Zheng, Chuanxia and Zhuang, Bohan and Pollefeys, Marc and Geiger, Andreas and Cham, Tat-Jen and Cai, Jianfei}.
\newblock Mvsplat: Efficient 3d gaussian splatting from sparse multi-view images.
\newblock \emph{European Conference on Computer Vision}, 2024.

\bibitem[Deitke et~al.(2023)Deitke, Schwenk, Salvador, Weihs, Michel, VanderBilt, Schmidt, Ehsani, Kembhavi, and Farhadi]{objaverse}
Matt Deitke, Dustin Schwenk, Jordi Salvador, Luca Weihs, Oscar Michel, Eli VanderBilt, Ludwig Schmidt, Kiana Ehsani, Aniruddha Kembhavi, and Ali Farhadi.
\newblock {Objaverse: A universe of annotated 3d objects}.
\newblock In \emph{Proceedings of the IEEE/CVF Conference on Computer Vision and Pattern Recognition (CVPR)}, pp.\  13142--13153, 2023.

\bibitem[Downs et~al.(2022)Downs, Francis, Koenig, Kinman, Hickman, Reymann, McHugh, and Vanhoucke]{gso}
Laura Downs, Anthony Francis, Nate Koenig, Brandon Kinman, Ryan Hickman, Krista Reymann, Thomas~B McHugh, and Vincent Vanhoucke.
\newblock {Google scanned objects: A high-quality dataset of 3d scanned household items}.
\newblock In \emph{2022 International Conference on Robotics and Automation (ICRA)}, pp.\  2553--2560. IEEE, 2022.

\bibitem[Forsyth \& Ponce(2003)Forsyth and Ponce]{pinhole1}
David~A Forsyth and Jean Ponce.
\newblock {A Modern Approach}.
\newblock \emph{Computer vision: a modern approach}, 17:\penalty0 21--48, 2003.

\bibitem[Guo et~al.(2022)Guo, Bautista, Colburn, Yang, Ulbricht, Susskind, and Shan]{FE-NVS}
Pengsheng Guo, Miguel~Angel Bautista, Alex Colburn, Liang Yang, Daniel Ulbricht, Joshua~M. Susskind, and Qi~Shan.
\newblock {Fast and Explicit Neural View Synthesis}.
\newblock In \emph{2022 IEEE/CVF Winter Conference on Applications of Computer Vision (WACV)}, Jan 2022.
\newblock \doi{10.1109/wacv51458.2022.00009}.
\newblock URL \url{http://dx.doi.org/10.1109/wacv51458.2022.00009}.

\bibitem[Hartley \& Zisserman(2003)Hartley and Zisserman]{pinhole2}
Richard Hartley and Andrew Zisserman.
\newblock \emph{{Multiple View Geometry in Computer Vision}}.
\newblock Cambridge university press, 2003.

\bibitem[Hong et~al.(2024)Hong, Zhang, Gu, Bi, Zhou, Liu, Liu, Sunkavalli, Bui, and Tan]{openlrm}
Yicong Hong, Kai Zhang, Jiuxiang Gu, Sai Bi, Yang Zhou, Difan Liu, Feng Liu, Kalyan Sunkavalli, Trung Bui, and Hao Tan.
\newblock {LRM: Large Reconstruction Model for Single Image to 3D}.
\newblock In \emph{International Conference on Learning Representations (ICLR)}, 2024.

\bibitem[Karras et~al.(2019)Karras, Laine, and Aila]{stylegan}
Tero Karras, Samuli Laine, and Timo Aila.
\newblock {A Style-Based Generator Architecture for Generative Adversarial Networks}.
\newblock In \emph{Proceedings of the IEEE/CVF conference on computer vision and pattern recognition (CVPR)}, pp.\  4401--4410, 2019.

\bibitem[Karras et~al.(2020)Karras, Laine, Aittala, Hellsten, Lehtinen, and Aila]{stylegan1}
Tero Karras, Samuli Laine, Miika Aittala, Janne Hellsten, Jaakko Lehtinen, and Timo Aila.
\newblock {Analyzing and Improving the Image Quality of StyleGAN}.
\newblock In \emph{Proceedings of the IEEE/CVF conference on computer vision and pattern recognition (CVPR)}, pp.\  8110--8119, 2020.

\bibitem[Kerbl et~al.(2023)Kerbl, Kopanas, Leimkuehler, and Drettakis]{3d-gs}
Bernhard Kerbl, Georgios Kopanas, Thomas Leimkuehler, and George Drettakis.
\newblock {3D Gaussian Splatting for Real-Time Radiance Field Rendering}.
\newblock In \emph{ACM Transactions on Graphics (TOG)}, 2023.

\bibitem[Kingma \& Ba(2015)Kingma and Ba]{adam}
Diederik Kingma and Jimmy Ba.
\newblock {Adam: A Method for Stochastic Optimization}.
\newblock In \emph{International Conference on Learning Representations (ICLR)}, San Diega, CA, USA, 2015.

\bibitem[Li et~al.(2023{\natexlab{a}})Li, Zhang, Zeng, Liu, Li, Ren, and Zhang]{dfa3d}
Hongyang Li, Hao Zhang, Zhaoyang Zeng, Shilong Liu, Feng Li, Tianhe Ren, and Lei Zhang.
\newblock {DFA3D: 3D Deformable Attention For 2D-to-3D Feature Lifting}.
\newblock In \emph{Proceedings of the IEEE/CVF international conference on computer vision (ICCV)}, 2023{\natexlab{a}}.

\bibitem[Li et~al.(2024)Li, Zhang, Liu, Zeng, Ren, Li, and Zhang]{taptr}
Hongyang Li, Hao Zhang, Shilong Liu, Zhaoyang Zeng, Tianhe Ren, Feng Li, and Lei Zhang.
\newblock {TAPTR: Tracking Any Point with Transformers as Detection}.
\newblock In \emph{Proceedings of the IEEE/CVF European Conference on Computer Vision (ECCV)}, 2024.

\bibitem[Li et~al.(2023{\natexlab{b}})Li, Tan, Zhang, Xu, Luan, Xu, Hong, Sunkavalli, Shakhnarovich, and Bi]{instant3d}
Jiahao Li, Hao Tan, Kai Zhang, Zexiang Xu, Fujun Luan, Yinghao Xu, Yicong Hong, Kalyan Sunkavalli, Greg Shakhnarovich, and Sai Bi.
\newblock {Instant3D: Fast Text-to-3D with Sparse-View Generation and Large Reconstruction Model}.
\newblock \emph{The International Conference on Learning Representations (ICLR)}, 2023{\natexlab{b}}.

\bibitem[Li et~al.(2022)Li, Wang, Li, Xie, Sima, Lu, Qiao, and Dai]{bevformer}
Zhiqi Li, Wenhai Wang, Hongyang Li, Enze Xie, Chonghao Sima, Tong Lu, Yu~Qiao, and Jifeng Dai.
\newblock {BEVFormer: Learning Bird’s-Eye-View Representation from Multi-Camera Images via Spatiotemporal Transformers}.
\newblock \emph{European conference on computer vision (ECCV)}, 2022.

\bibitem[Lin et~al.(2022)Lin, Lin, Lai, Lin, Shih, and Ramamoorthi]{visionnerf}
Kai-En Lin, Yen-Chen Lin, Wei-Sheng Lai, Tsung-Yi Lin, Yichang Shih, and Ravi Ramamoorthi.
\newblock {Vision Transformer for NeRF-Based View Synthesis from a Single Input Image}.
\newblock In \emph{2023 IEEE/CVF Winter Conference on Applications of Computer Vision (WACV)}, 2022.

\bibitem[Liu et~al.(2024{\natexlab{a}})Liu, Jin, Zeng, Han, and Zhang]{liu2024comprehensive}
Kenkun Liu, Derong Jin, Ailing Zeng, Xiaoguang Han, and Lei Zhang.
\newblock {A Comprehensive Benchmark for Neural Human Radiance Fields}.
\newblock \emph{Advances in Neural Information Processing Systems (NIPS)}, 36, 2024{\natexlab{a}}.

\bibitem[Liu et~al.(2024{\natexlab{b}})Liu, Shi, Chen, Zhang, Xu, Wei, Chen, Zeng, Gu, and Su]{One-2-3-45++}
Minghua Liu, Ruoxi Shi, Linghao Chen, Zhuoyang Zhang, Chao Xu, Xinyue Wei, Hansheng Chen, Chong Zeng, Jiayuan Gu, and Hao Su.
\newblock {One-2-3-45++: Fast Single Image to 3D Objects with Consistent Multi-View Generation and 3D Diffusion}.
\newblock In \emph{Proceedings of the IEEE/CVF Conference on Computer Vision and Pattern Recognition}, pp.\  10072--10083, 2024{\natexlab{b}}.

\bibitem[Liu et~al.(2024{\natexlab{c}})Liu, Xu, Jin, Chen, Varma~T, Xu, and Su]{one2345}
Minghua Liu, Chao Xu, Haian Jin, Linghao Chen, Mukund Varma~T, Zexiang Xu, and Hao Su.
\newblock {One-2-3-45: Any single image to 3d mesh in 45 seconds without per-shape optimization}.
\newblock In \emph{Conference on Neural Information Processing Systems (NIPS)}, 2024{\natexlab{c}}.

\bibitem[Liu et~al.(2024{\natexlab{d}})Liu, Zeng, Wei, Shi, Chen, Xu, Zhang, Wang, Zhang, Liu, Wu, and Su]{meshformer}
Minghua Liu, Chong Zeng, Xinyue Wei, Ruoxi Shi, Linghao Chen, Chao Xu, Mengqi Zhang, Zhaoning Wang, Xiaoshuai Zhang, Isabella Liu, Hongzhi Wu, and Hao Su.
\newblock {MeshFormer: High-Quality Mesh Generation with 3D-Guided Reconstruction Model}.
\newblock \emph{Conference on Neural Information Processing Systems (NIPS)}, 2024{\natexlab{d}}.

\bibitem[Liu et~al.(2023{\natexlab{a}})Liu, Wu, Hoorick, Tokmakov, Zakharov, and Vondrick]{zero123}
Ruoshi Liu, Rundi Wu, Basile~Van Hoorick, Pavel Tokmakov, Sergey Zakharov, and Carl Vondrick.
\newblock {Zero-1-to-3: Zero-shot One Image to 3D Object}.
\newblock In \emph{IEEE/CVF International Conference on Computer Vision (ICCV)}, 2023{\natexlab{a}}.

\bibitem[Liu et~al.(2022)Liu, Li, Zhang, Yang, Qi, Su, Zhu, and Zhang]{dab-detr}
Shilong Liu, Feng Li, Hao Zhang, Xiao Yang, Xianbiao Qi, Hang Su, Jun Zhu, and Lei Zhang.
\newblock {DAB}-{DETR}: Dynamic anchor boxes are better queries for {DETR}.
\newblock In \emph{International Conference on Learning Representations (ICLR)}, 2022.

\bibitem[Liu et~al.(2023{\natexlab{b}})Liu, Ren, Chen, Zeng, Zhang, Li, Li, Huang, Su, Zhu, and Zhang]{detr-matching}
Siyi Liu, Tianhe Ren, Jia-Yu Chen, Zhaoyang Zeng, Hao Zhang, Feng Li, Hongyang Li, Jun Huang, Hang Su, Jun-Juan Zhu, and Lei Zhang.
\newblock {Stable-DINO: Detection Transformer with Stable Matching}.
\newblock In \emph{IEEE/CVF International Conference on Computer Vision (ICCV)}, 2023{\natexlab{b}}.

\bibitem[Liu et~al.(2021)Liu, Lin, Cao, Hu, Wei, Zhang, Lin, and Guo]{swinT}
Ze~Liu, Yutong Lin, Yue Cao, Han Hu, Yixuan Wei, Zheng Zhang, Stephen Lin, and Baining Guo.
\newblock {Swin Transformer: Hierarchical vision transformer using shifted windows}.
\newblock In \emph{Proceedings of the IEEE/CVF international conference on computer vision (ICCV)}, pp.\  10012--10022, 2021.

\bibitem[Mildenhall et~al.(2020)Mildenhall, Srinivasan, Tancik, Barron, Ramamoorthi, and Ng]{nerf}
Ben Mildenhall, Pratul~P. Srinivasan, Matthew Tancik, Jonathan~T. Barron, Ravi Ramamoorthi, and Ren Ng.
\newblock {NeRF: Representing Scenes as Neural Radiance Fields for View Synthesis}.
\newblock In \emph{The European Conference on Computer Vision (ECCV)}, 2020.

\bibitem[M{\"u}ller et~al.(2022)M{\"u}ller, Evans, Schied, and Keller]{instantnerf}
Thomas M{\"u}ller, Alex Evans, Christoph Schied, and Alexander Keller.
\newblock {Instant Neural Graphics Primitives with a Multiresolution Hash Encoding}.
\newblock In \emph{ACM Transactions on Graphics (SIGGRAPH)}, 2022.

\bibitem[Narang et~al.(2018)Narang, Diamos, Elsen, Micikevicius, Alben, Garcia, Ginsburg, Houston, Kuchaiev, Venkatesh, et~al.]{mixed_pres}
Sharan Narang, Gregory Diamos, Erich Elsen, Paulius Micikevicius, Jonah Alben, David Garcia, Boris Ginsburg, Michael Houston, Oleksii Kuchaiev, Ganesh Venkatesh, et~al.
\newblock {Mixed Precision Training}.
\newblock In \emph{6th international conference on learning representations (ICLR)}, volume~1, pp.\ ~14, 2018.

\bibitem[Qi et~al.(2017{\natexlab{a}})Qi, Su, Mo, and Guibas]{pointnet}
Charles~R Qi, Hao Su, Kaichun Mo, and Leonidas~J Guibas.
\newblock {PointNet: Deep Learning on Point Sets for 3D Classification and Segmentation}.
\newblock In \emph{Proceedings of the IEEE conference on computer vision and pattern recognition (CVPR)}, pp.\  652--660, 2017{\natexlab{a}}.

\bibitem[Qi et~al.(2017{\natexlab{b}})Qi, Yi, Su, and Guibas]{pointnet++}
Charles~Ruizhongtai Qi, Li~Yi, Hao Su, and Leonidas~J Guibas.
\newblock {PointNet++: Deep Hierarchical Feature Learning on Point Sets in a Metric Space}.
\newblock \emph{Advances in neural information processing systems (NIPS)}, 30, 2017{\natexlab{b}}.

\bibitem[Rombach et~al.(2022)Rombach, Blattmann, Lorenz, Esser, and Ommer]{stablediffusion}
Robin Rombach, Andreas Blattmann, Dominik Lorenz, Patrick Esser, and Bj{\"o}rn Ommer.
\newblock {High-resolution image synthesis with latent diffusion models}.
\newblock In \emph{Proceedings of the IEEE/CVF conference on computer vision and pattern recognition (CVPR)}, pp.\  10684--10695, 2022.

\bibitem[Ronneberger et~al.(2015)Ronneberger, Fischer, and Brox]{unet}
Olaf Ronneberger, Philipp Fischer, and Thomas Brox.
\newblock {U-Net: Convolutional Networks for Biomedical Image Segmentation}.
\newblock In \emph{Medical image computing and computer-assisted intervention (MICCAI)}, 2015.

\bibitem[Shen et~al.(2024)Shen, Wu, Yi, Zhou, Zhang, Yan, and Wang]{gamba}
Qiuhong Shen, Zike Wu, Xuanyu Yi, Pan Zhou, Hanwang Zhang, Shuicheng Yan, and Xinchao Wang.
\newblock {Gamba: Marry gaussian splatting with mamba for single view 3d reconstruction}.
\newblock \emph{arXiv preprint arXiv:2403.18795}, 2024.

\bibitem[Shi et~al.(2024)Shi, Wang, Ye, Mai, Li, and Yang]{mvdream}
Yichun Shi, Peng Wang, Jianglong Ye, Long Mai, Kejie Li, and Xiao Yang.
\newblock {MVDream: Multi-view Diffusion for 3D Generation}.
\newblock \emph{The International Conference on Learning Representations (ICLR)}, 2024.

\bibitem[Song et~al.(2021{\natexlab{a}})Song, Meng, and Ermon]{ddim}
Jiaming Song, Chenlin Meng, and Stefano Ermon.
\newblock {Denoising Diffusion Implicit Models}.
\newblock In \emph{International Conference on Learning Representations (ICLR)}, 2021{\natexlab{a}}.

\bibitem[Song et~al.(2021{\natexlab{b}})Song, Sohl-Dickstein, Kingma, Kumar, Ermon, and Poole]{songunet}
Yang Song, Jascha Sohl-Dickstein, Diederik~P Kingma, Abhishek Kumar, Stefano Ermon, and Ben Poole.
\newblock {Score-Based Generative Modeling through Stochastic Differential Equations}.
\newblock \emph{The International Conference on Learning Representations (ICLR)}, 2021{\natexlab{b}}.
\newblock URL \url{https://openreview.net/forum?id=PxTIG12RRHS}.

\bibitem[Szymanowicz et~al.(2024)Szymanowicz, Rupprecht, and Vedaldi]{SplatterImage}
Stanislaw Szymanowicz, Christian Rupprecht, and Andrea Vedaldi.
\newblock {Splatter Image: Ultra-Fast Single-View 3D Reconstruction}.
\newblock In \emph{IEEE/CVF Conference on Computer Vision and Pattern Recognition (CVPR)}, 2024.

\bibitem[Tang et~al.(2024{\natexlab{a}})Tang, Chen, Chen, Wang, Zeng, and Liu]{LGM}
Jiaxiang Tang, Zhaoxi Chen, Xiaokang Chen, Tengfei Wang, Gang Zeng, and Ziwei Liu.
\newblock {LGM: Large Multi-View Gaussian Model for High-Resolution 3D Content Creation}.
\newblock In \emph{European Conference on Computer Vision}, pp.\  1--18. Springer, 2024{\natexlab{a}}.

\bibitem[Tang et~al.(2024{\natexlab{b}})Tang, Ren, Zhou, Liu, and Zeng]{dreamgaussian}
Jiaxiang Tang, Jiawei Ren, Hang Zhou, Ziwei Liu, and Gang Zeng.
\newblock {DreamGaussian: Generative Gaussian Splatting for Efficient 3D Content Creation}.
\newblock In \emph{International Conference on Learning Representations (ICLR)}, 2024{\natexlab{b}}.

\bibitem[Tochilkin et~al.(2024)Tochilkin, Pankratz, Liu, Huang, , Letts, Li, Liang, Laforte, Jampani, and Cao]{TripoSR}
Dmitry Tochilkin, David Pankratz, Zexiang Liu, Zixuan Huang, , Adam Letts, Yangguang Li, Ding Liang, Christian Laforte, Varun Jampani, and Yan-Pei Cao.
\newblock {TripoSR: Fast 3D Object Reconstruction from a Single Image}.
\newblock \emph{arXiv preprint arXiv:2403.02151}, 2024.

\bibitem[Viazovetskyi et~al.(2020)Viazovetskyi, Ivashkin, and Kashin]{stylegan2}
Yuri Viazovetskyi, Vladimir Ivashkin, and Evgeny Kashin.
\newblock {StyleGAN2 Distillation for Feed-forward Image Manipulation}.
\newblock In \emph{Proceedings of the IEEE/CVF European Conference on Computer Vision (ECCV)}, pp.\  170--186. Springer, 2020.

\bibitem[Wang \& Shi(2023)Wang and Shi]{imagedream}
Peng Wang and Yichun Shi.
\newblock {ImageDream: Image-Prompt Multi-view Diffusion for 3D Generation}.
\newblock \emph{arXiv preprint arXiv:2312.02201}, 2023.

\bibitem[Wang et~al.(2021{\natexlab{a}})Wang, Wang, Genova, Srinivasan, Zhou, Barron, Martin-Brualla, Snavely, and Funkhouser]{IBRNet}
Qianqian Wang, Zhicheng Wang, Kyle Genova, Pratul Srinivasan, Howard Zhou, Jonathan~T. Barron, Ricardo Martin-Brualla, Noah Snavely, and Thomas Funkhouser.
\newblock {IBRNet: Learning Multi-View Image-Based Rendering}.
\newblock In \emph{2021 IEEE/CVF Conference on Computer Vision and Pattern Recognition (CVPR)}, Jun 2021{\natexlab{a}}.
\newblock \doi{10.1109/cvpr46437.2021.00466}.
\newblock URL \url{http://dx.doi.org/10.1109/cvpr46437.2021.00466}.

\bibitem[Wang et~al.(2021{\natexlab{b}})Wang, Xie, Li, Fan, Song, Liang, Lu, Luo, and Shao]{pvt}
Wenhai Wang, Enze Xie, Xiang Li, Deng-Ping Fan, Kaitao Song, Ding Liang, Tong Lu, Ping Luo, and Ling Shao.
\newblock {Pyramid vision transformer: A versatile backbone for dense prediction without convolutions}.
\newblock In \emph{Proceedings of the IEEE/CVF international conference on computer vision (ICCV)}, pp.\  568--578, 2021{\natexlab{b}}.

\bibitem[Wang et~al.(2004)Wang, Bovik, Sheikh, and Simoncelli]{ssim}
Zhou Wang, Alan~C Bovik, Hamid~R Sheikh, and Eero~P Simoncelli.
\newblock {Image Quality Assessment: From Error Visibility to Structural Similarity}.
\newblock \emph{IEEE transactions on image processing}, 13\penalty0 (4):\penalty0 600--612, 2004.

\bibitem[Wei et~al.(2024)Wei, Zhang, Bi, Tan, Luan, Deschaintre, Sunkavalli, Su, and Xu]{meshlrm}
Xinyue Wei, Kai Zhang, Sai Bi, Hao Tan, Fujun Luan, Valentin Deschaintre, Kalyan Sunkavalli, Hao Su, and Zexiang Xu.
\newblock {MeshLRM: Large Reconstruction Model for High-Quality Mesh}.
\newblock \emph{arXiv preprint arXiv:2404.12385}, 2024.

\bibitem[Wu et~al.(2025)Wu, Liu, Gao, Jiang, and Zhang]{wu2024dig3d}
Jiamin Wu, Kenkun Liu, Han Gao, Xiaoke Jiang, and Lei Zhang.
\newblock {LeanGaussian: Breaking Pixel or Point Cloud Correspondence in Modeling 3D Gaussians}.
\newblock \emph{Conference on Computer Vision and Pattern Recognition (CVPR)}, 2025.

\bibitem[Xiong et~al.(2024)Xiong, Li, Liu, Liao, Hu, Zhu, Ning, Qiu, Wang, Wang, et~al.]{xiong2024mvhumannet}
Zhangyang Xiong, Chenghong Li, Kenkun Liu, Hongjie Liao, Jianqiao Hu, Junyi Zhu, Shuliang Ning, Lingteng Qiu, Chongjie Wang, Shijie Wang, et~al.
\newblock {MVHumanNet: A Large-scale Dataset of Multi-view Daily Dressing Human Captures}.
\newblock In \emph{Proceedings of the IEEE/CVF Conference on Computer Vision and Pattern Recognition (CVPR)}, pp.\  19801--19811, 2024.

\bibitem[Xu et~al.(2024{\natexlab{a}})Xu, Cheng, Gao, Wang, Gao, and Shan]{instantmesh}
Jiale Xu, Weihao Cheng, Yiming Gao, Xintao Wang, Shenghua Gao, and Ying Shan.
\newblock {InstantMesh: Efficient 3D Mesh Generation from a Single Image with Sparse-view Large Reconstruction Models}.
\newblock \emph{arXiv preprint arXiv:2404.07191}, 2024{\natexlab{a}}.

\bibitem[Xu et~al.(2024{\natexlab{b}})Xu, Shi, Yifan, Peng, Yang, Shen, and Gordon]{grm}
Yinghao Xu, Zifan Shi, Wang Yifan, Sida Peng, Ceyuan Yang, Yujun Shen, and Wetzstein Gordon.
\newblock {GRM: Large Gaussian Reconstruction Model for Efficient 3D Reconstruction and Generation}, author={Xu Yinghao and Shi Zifan and Yifan Wang and Chen Hansheng and Yang Ceyuan and Peng Sida and Shen Yujun and Wetzstein Gordon}, 2024{\natexlab{b}}.

\bibitem[Yi et~al.(2024)Yi, Wu, Shen, Xu, Zhou, Lim, Yan, Wang, and Zhang]{mvgamba}
Xuanyu Yi, Zike Wu, Qiuhong Shen, Qingshan Xu, Pan Zhou, Joo-Hwee Lim, Shuicheng Yan, Xinchao Wang, and Hanwang Zhang.
\newblock {MVGamba: Unify 3D Content Generation as State Space Sequence Modeling}.
\newblock \emph{arXiv preprint arXiv:2406.06367}, 2024.

\bibitem[Yu et~al.(2021)Yu, Ye, Tancik, and Kanazawa]{pixelnerf}
Alex Yu, Vickie Ye, Matthew Tancik, and Angjoo Kanazawa.
\newblock {pixelNeRF: Neural Radiance Fields from One or Few Images}.
\newblock In \emph{IEEE/CVF Conference on Computer Vision and Pattern Recognition (CVPR)}, 2021.

\bibitem[Zhang et~al.(2024{\natexlab{a}})Zhang, Song, Wei, Chen, Lu, and Tang]{geolrm}
Chubin Zhang, Hongliang Song, Yi~Wei, Yu~Chen, Jiwen Lu, and Yansong Tang.
\newblock {GeoLRM: Geometry-Aware Large Reconstruction Model for High-Quality 3D Gaussian Generation}.
\newblock \emph{NeurIPS}, 2024{\natexlab{a}}.

\bibitem[Zhang et~al.(2023)Zhang, Li, Liu, Zhang, Su, Zhu, shuan Ni, and yeung Shum]{dino}
Hao Zhang, Feng Li, Shilong Liu, Lei Zhang, Hang Su, Jun-Juan Zhu, Lionel~Ming shuan Ni, and Heung yeung Shum.
\newblock {DINO: DETR with Improved DeNoising Anchor Boxes for End-to-End Object Detection}.
\newblock In \emph{The International Conference on Learning Representations (ICLR)}, 2023.

\bibitem[Zhang et~al.(2024{\natexlab{b}})Zhang, Bi, Tan, Xiangli, Zhao, Sunkavalli, and Xu]{gslrm}
Kai Zhang, Sai Bi, Hao Tan, Yuanbo Xiangli, Nanxuan Zhao, Kalyan Sunkavalli, and Zexiang Xu.
\newblock {GS-LRM: Large Reconstruction Model for 3D Gaussian Splatting}.
\newblock \emph{European Conference on Computer Vision}, 2024{\natexlab{b}}.

\bibitem[Zhang et~al.(2018{\natexlab{a}})Zhang, Isola, Efros, Shechtman, and Wang]{LIPIS}
Richard Zhang, Phillip Isola, Alexei~A. Efros, Eli Shechtman, and Oliver Wang.
\newblock {The Unreasonable Effectiveness of Deep Features as a Perceptual Metric}.
\newblock In \emph{IEEE/CVF Conference on Computer Vision and Pattern Recognition (CVPR)}, 2018{\natexlab{a}}.

\bibitem[Zhang et~al.(2018{\natexlab{b}})Zhang, Isola, Efros, Shechtman, and Wang]{lpips}
Richard Zhang, Phillip Isola, Alexei~A Efros, Eli Shechtman, and Oliver Wang.
\newblock {The Unreasonable Effectiveness of Deep Features as a Perceptual Metric}.
\newblock In \emph{Proceedings of the IEEE conference on computer vision and pattern recognition (CVPR)}, pp.\  586--595, 2018{\natexlab{b}}.

\bibitem[Zheng \& Vedaldi(2024)Zheng and Vedaldi]{free3d}
Chuanxia Zheng and Andrea Vedaldi.
\newblock {Free3D: Consistent Novel View Synthesis without 3D Representation}.
\newblock In \emph{IEEE/CVF Conference on Computer Vision and Pattern Recognition (CVPR)}, 2024.

\bibitem[Zhu et~al.(2021)Zhu, Su, Lu, Li, Wang, and Dai]{DETR}
Xizhou Zhu, Weijie Su, Lewei Lu, Bin Li, Xiaogang Wang, and Jifeng Dai.
\newblock {Deformable DETR: Deformable Transformers for End-to-End Object Detection}.
\newblock In \emph{The International Conference on Learning Representations (ICLR)}, 2021.

\bibitem[Zou et~al.(2024)Zou, Yu, Guo, Li, Liang, Cao, and Zhang]{triplane-gs}
Zi-Xin Zou, Zhipeng Yu, Yuan-Chen Guo, Yangguang Li, Ding Liang, Yan-Pei Cao, and Song-Hai Zhang.
\newblock {Triplane Meets Gaussian Splatting: Fast and Generalizable Single-View 3D Reconstruction with Transformers}.
\newblock In \emph{Proceedings of the IEEE/CVF Conference on Computer Vision and Pattern Recognition (CVPR)}, pp.\  10324--10335, June 2024.

\bibitem[Zwicker et~al.(2001)Zwicker, Pfister, van Baar, and Gross]{EWASplatting}
M.~Zwicker, H.~Pfister, J.~van Baar, and M.~Gross.
\newblock {EWA volume splatting}.
\newblock In \emph{IEEE Visualization (IEEE VIS)}, 2001.

\end{thebibliography}
\bibliographystyle{iclr2025_conference}

\newpage
\appendix
\section{Appendix}
\subsection{Model details}

\subsubsection{Spatial efficient self attention (SESA)} \label{app: space_efficient_self_attn}
\begin{figure}
    \centering
    \includegraphics[width=0.35\textwidth]{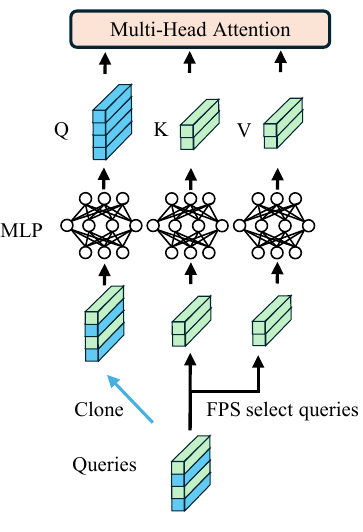}
    \caption{Spatially Efficient Self-Attention: While employing all queries as query in the self-attention mechanism, we leverage Farthest Point Sampling (FPS) to downsample certain 3D Gaussians. This process enables the extraction of their corresponding queries as keys and values within the self-attention operation.}
    \label{fig: self_attn}
\end{figure}

While our 3D-aware deformable attention mechanism is notably efficient, the computational cost and memory occupation mainly arises in the self-attention component, particularly when dealing with a large number of 3D Gaussians. However, updating each 3D Gaussian with information from all others is not always necessary because those neighbouring 3D Gaussians usually carry similar information. 

To mitigate this issue, as depicted in \cref{fig: self_attn} and drawing inspiration from \cite{pvt}, we introduce a technique aimed at reducing the size of the key and value components while maintaining the query component unaltered within the self-attention process. The core insights behind this approach is that while each 3D Gaussian requires updating, not every other 3D Gaussian needs to contribute to this update. We achieve this by selectively updating each query solely with a subset of corresponding queries linked to other 3D Gaussians. 

More specifically, we leverage the Fast Point Sampling (FPS) algorithm commonly used in point cloud methodologies like PointNet \citep{pointnet} and PointNet++ \citep{pointnet++}. We employ the Gaussian centers $\mu$ to identify the most distantly located points and use these points to index the queries. By implementing this strategy, our model significantly reduces the model's overall memory footprint while preserving essential information exchange among the Gaussians. 

\subsubsection{Analysis for the regression of the center of 3D Gaussians and the initialization for 3D Gaussians and queries} \label{app: analysis}

\paragraph{From camera space to world space}
In \cite{SplatterImage}, the Gaussian centers are located in each input view's camera space as shown in \cref{eq: gaussian_center},
\begin{equation} \label{eq: gaussian_center}
    \boldsymbol{\boldsymbol{\mu}}_{cam} = 
    \begin{bmatrix}
        x_{cam}\\
        y_{cam} \\
        z_{cam} \\
    \end{bmatrix}
    =
    \begin{bmatrix}
        u_1d + \Delta_{x}\\
        u_2d + \Delta_{y} \\
        d + \Delta_{z}
    \end{bmatrix}
\end{equation}
The center coordinates $x_{\text{cam}}, y_{\text{cam}}, z_{\text{cam}}$ are parameterized by depth $d$ and offsets $(\Delta_{x}, \Delta_{y}, \Delta_{z})$. $u_1, u_2$ are the UV coordinates of the ray passing through the corresponding input image. This design represents each point with multiple Gaussians, potentially introducing `ghosting' due to concatenation issues at various points caused by depth inaccuracies and tend to shortcut input views \citep{wu2024dig3d}. In our framework, we define unitary Gaussians in world space, project their centers to each input view for feature retrieval, as depicted in \cref{fig: mvdfa}(a). The centers of Gaussians can be written as $\boldsymbol{\mu}_{\text{world}}=[x_{\text{world}}, y_{\text{world}}, z_{\text{world}}]$. 

\paragraph{3D Gaussian initialization}
However, during the initial training phases, discrepancies between the 3D Gaussian centers and ground truth often result in imprecise selection of image features at sampling points, presenting challenges for model convergence as shown in \cref{sec: ablation}.
To address this issue, we utilize a initialization that directly regress 3D Gaussian parameters from each pixel of the image features. We first train a coarse network that predicts 3D Gaussians for each pixel of each input view under the camera space of that view. After that, the 3D Gaussians are transformed and fused to get the 3D Gaussians under world space. The role of this network is to provide a coarse initialization of 3D Gaussians for the subsequent refinement.
We use the UNet architecture, which has the same structure to the feature extractor. 

Moreover, We employ a relative coordinate system, where the camera poses for all views are known. The initial input view is established as the world coordinates (with the camera pose represented by the identity matrix), and subsequently, all other views are transformed to align with the reference view. This approach allows us to represent all 3D data within this consistent relative coordinate system.

\subsection{Experiment details} 
\subsubsection{Datasets} \label{app: datasets}
We utilized a refined subset of the Objaverse LVIS dataset \citep{objaverse} for both training and validating our model. This subset was curated to exclude low-quality models, resulting in a dataset containing 36,044 high-quality objects. This open-category dataset encompasses a diverse range of objects commonly encountered in everyday scenarios. For training, we leveraged rendered images provided by zero-1-to-3 \citep{zero123} for the random input setting. Each object in the dataset is associated with approximately 12 random views, accompanied by their respective camera poses. We partitioned 99\% of the objects for training purposes, reserving the remaining 1\% for validation. During training, we randomly selected a subset of views as input while using all 12 views for supervision. Each rendered image has a resolution of $512 \times 512$, which we downscaled to $128 \times 128$. For the fixed view setting, we render the images with fixed views as input and 32 more random views with elevation in $(-30, 30)$ degrees for supervision.

To evaluate our model's performance in open-category settings, we conducted tests on the Google Scanned Objects (GSO) benchmark \citep{gso}. The GSO dataset comprises 1,030 3D objects categorized into 17 classes. For this evaluation, we utilized rendered images sourced from Free3D \citep{free3d}, which consist of 25 random views along with their corresponding camera poses. Notably, there are no restrictions on the elevation of the rendered views. We utilized the initial views as inputs and the remaining views for assessing our novel view synthesis task. Additionally, we observed that LGM \citep{LGM} only support fixed-view inputs (e.g., front, left, back, and right). To address this, we evaluated a new rendered GSO dataset at 0 degrees elevation, testing it on 32 random views with elevations ranging from 0 to 30 degrees. To distinguish between the two test sets, we refer to them as GSO and GSO respectively in the following analysis. 

\subsubsection{Implementation details} \label{sec: exp_setting}
We train our model on the setting of 4 views, each time we randomly select 4 views as input and all the views for supervision. For the initialization, we train the model with less views (i.e. 2 views) with resolution $128 \times 128$ and generate 16384 3D Gaussians as initialization of the fine stage. In the fine stage, We use 19600 3D Gaussians to represent the 3D object. For the 3D Gaussians from the initialization, we use the mask to remove the background points and padding the number of 3D Gaussians to 19600 by copying some of the remaining 3D Gaussians. The selected 3D Gaussians are then utilized to project queries onto image plane in the refine stage. In each deformable attention layer, we utilize 4 sampling points for each projected 3D Gaussian reference point to sample values on the image.

We use 4 decoder layers and the hidden dimension is 256. We use a mixed-precision training \citep{mixed_pres} with BF16 data type. We train our model with Adam \citep{adam} optimizer and the learning rate is 0.0001. We take 300K iteration with batch size 4.
For the initialization, we train it on 8 3090 GPUs (24G) for 5 days and for the refinement stage, we train it on 8 A100 (80G) for 3 days. 

\subsection{More results} \label{app: more_result}
\subsubsection{Input views with random camera poses} \label{app: random_view}
\begin{table}
  \footnotesize
  \caption{Quantitative results for inputting 4 views on GSO dataset.}
  \label{tab: 4_random_view_results}
  \centering
  \begin{tabular}{llll}
    \toprule
       Method & PSNR $\uparrow$     & SSIM $\uparrow$ & LPIPS $\downarrow$ \\
    \midrule
    Splatter Image & 25.7660 & 0.8932 & 0.2575 \\
    LGM & 15.1113 &  0.8440 &  0.1592 \\
    InstantMesh & 17.3073 & 0.8525 & 0.1376 \\ 
    \midrule
    Our Model & \textbf{26.3020} & \textbf{0.9255} & \textbf{0.0836}  \\
    \bottomrule
  \end{tabular}
\end{table}
Previous methods (LGM and InstantMesh) usually rely on fixed views as input, as they align well with views generated from diffusion models like ImageDream \citep{imagedream}. In real-world scenarios, users are more inclined to provide random views as input. \cref{tab: 4_random_view_results} displays the results when utilizing random 4 views as input on the GSO dataset. Notably, there is a performance drop observed in LGM and InstantMesh with random input views. For Splatter Image, although the PSNR does not reduced much, its SSIM and LPIPS reduced significantly. We provide more visualization in \cref{fig: fix_vs_rand}.

\begin{figure*}
    \centering
    \includegraphics[width=1.0\textwidth]{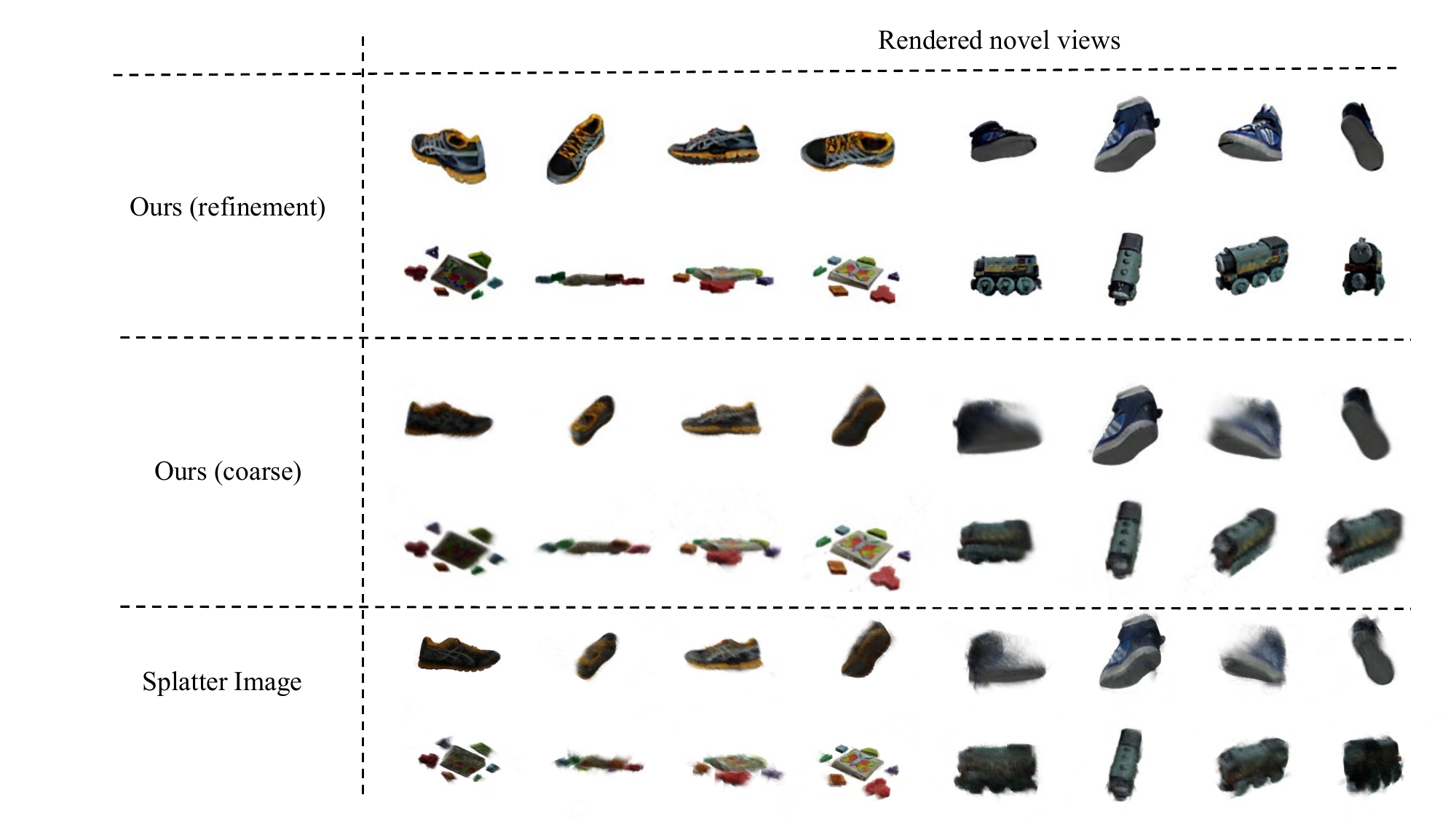}
    \caption{Visualization for Splatter Image with fixed view input and random view input.}
    \label{fig: random_vis_tab2}
\end{figure*}

PSNR of Splatter Image in \cref{tab: 4_random_view_results} is good but SSIM and LPIPS are not good enough, we further provide the visualization is in \cref{fig: random_vis_tab2}.
\subsubsection{More example for `Ghosting' problem} \label{app: ghosting}

\begin{figure*}
    \centering
    \includegraphics[width=1.0\textwidth]{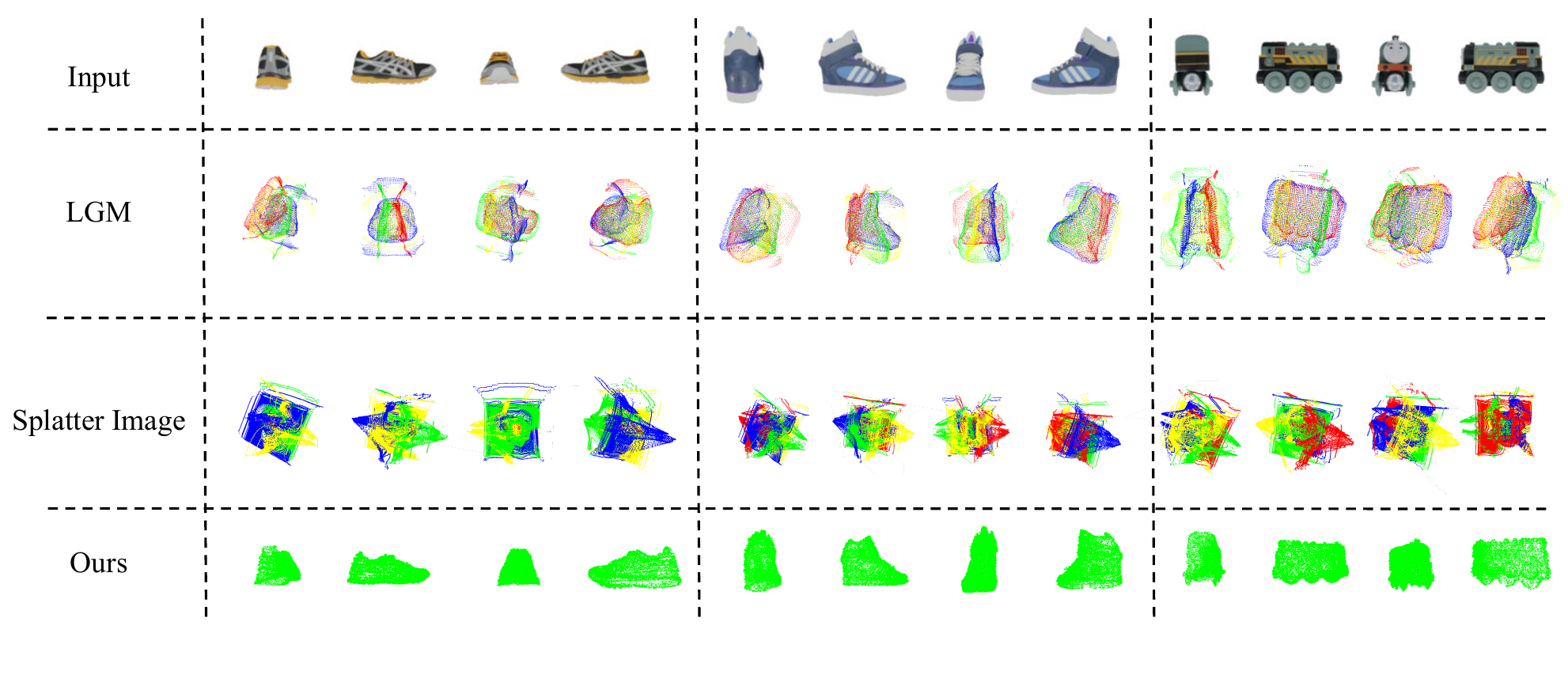}
    \caption{Point clouds of the center of Gaussians from each view. The Gaussians from different views are in different colors.}
    \label{fig: view_point_cloud}
\end{figure*}

We gives more `ghosting' visualization problem by visualize center of Gaussians from each view in different colors, as shown in \cref{fig: view_point_cloud}. Gaussians from different views representing the same part of the object may lays on the different position in the 3D space and thus cause the `ghosting' problem.

\subsubsection{More visualization} \label{app: more_vis}

Image-to-3D conversion represents a fundamental application in 3D generation. Following the methodology of LGM and InstantMesh \citep{LGM, instantmesh}, we first leverage a multi-view diffusion model, ImageDream \citep{imagedream}, to generate four predetermined views. Subsequently, our model is employed for 3D Gaussian reconstruction. A comparative analysis with LGM and InstantMesh is detailed in \cref{tab: single_view_result}. For this particular scenario, we utilize the fixed-view GSO test set with elevations ranging between 0 and 30 degrees. Given potential variations in camera poses among the generated multi-views, which may not align precisely with standard front, right, back, and left perspectives, we selectively retain 266 objects that consistently yield accurate images under the provided camera poses. 

\begin{table*}[htp]
  \caption{Quantitative results for single view reconstruction on GSO dataset.}
  \label{tab: single_view_result}
  \centering
  \begin{tabular}{llll}
    \toprule
       Method & PSNR $\uparrow$     & SSIM $\uparrow$ & LPIPS $\downarrow$ \\
    \midrule
    LGM \citep{LGM} & 20.8139 & 0.8581 & 0.1508 \\
    InstantMesh \citep{instantmesh} & 19.4667 & 0.8379 & 0.1842 \\ 
    \midrule
    Our Model & \textbf{22.3534} & \textbf{0.8567} & \textbf{0.1492} \\
    \bottomrule
  \end{tabular}
\end{table*}

\begin{figure}
    \centering
    \includegraphics[width=0.98\textwidth]{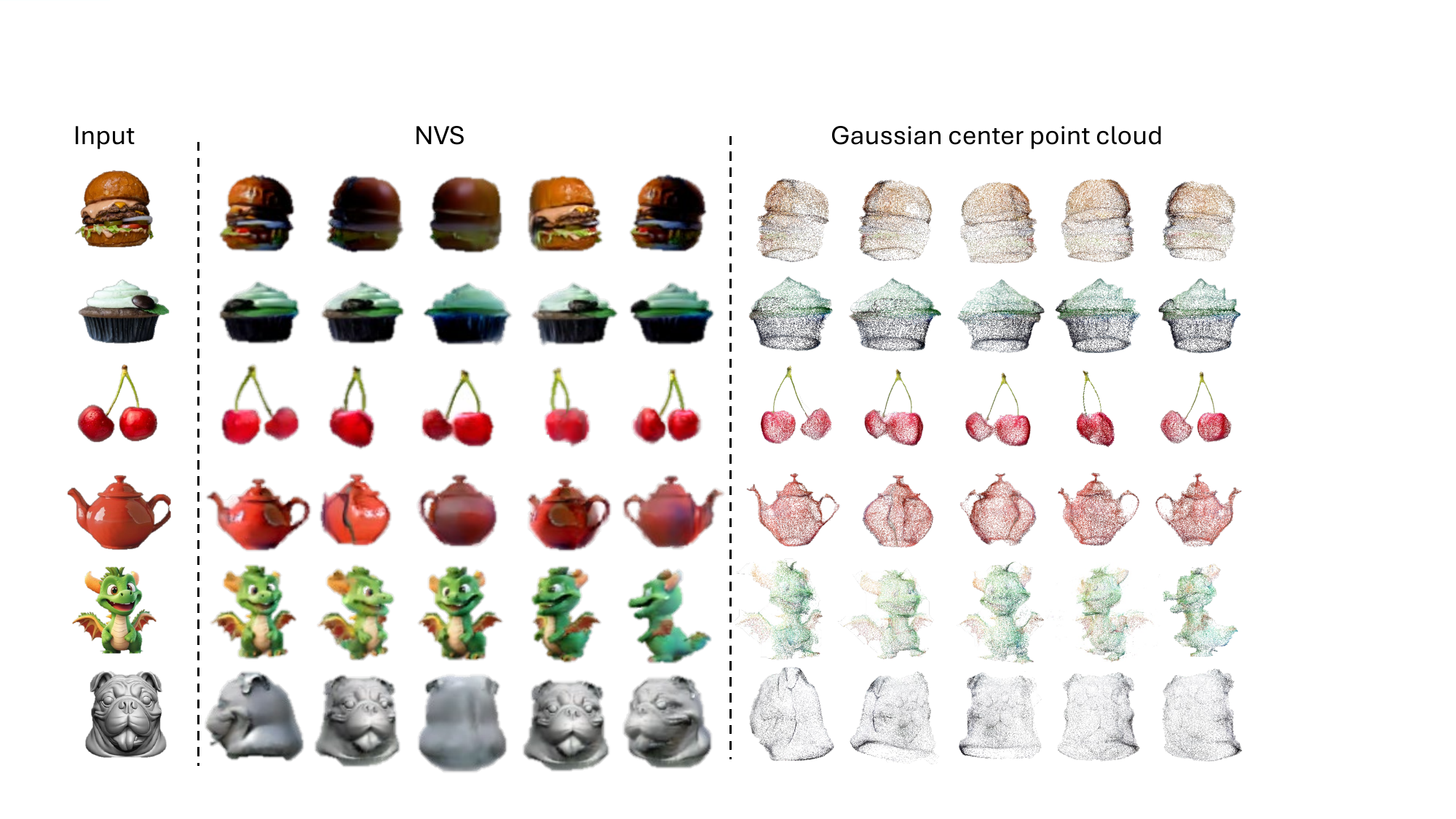}
    \caption{Quality for rendered novel views on in the wild data for inputting 1 view and using ImageDream to generate 4 views.}
    \label{fig: in_the_wild}
\end{figure}

\begin{figure*}
    \centering
    \includegraphics[width=1.0\textwidth]{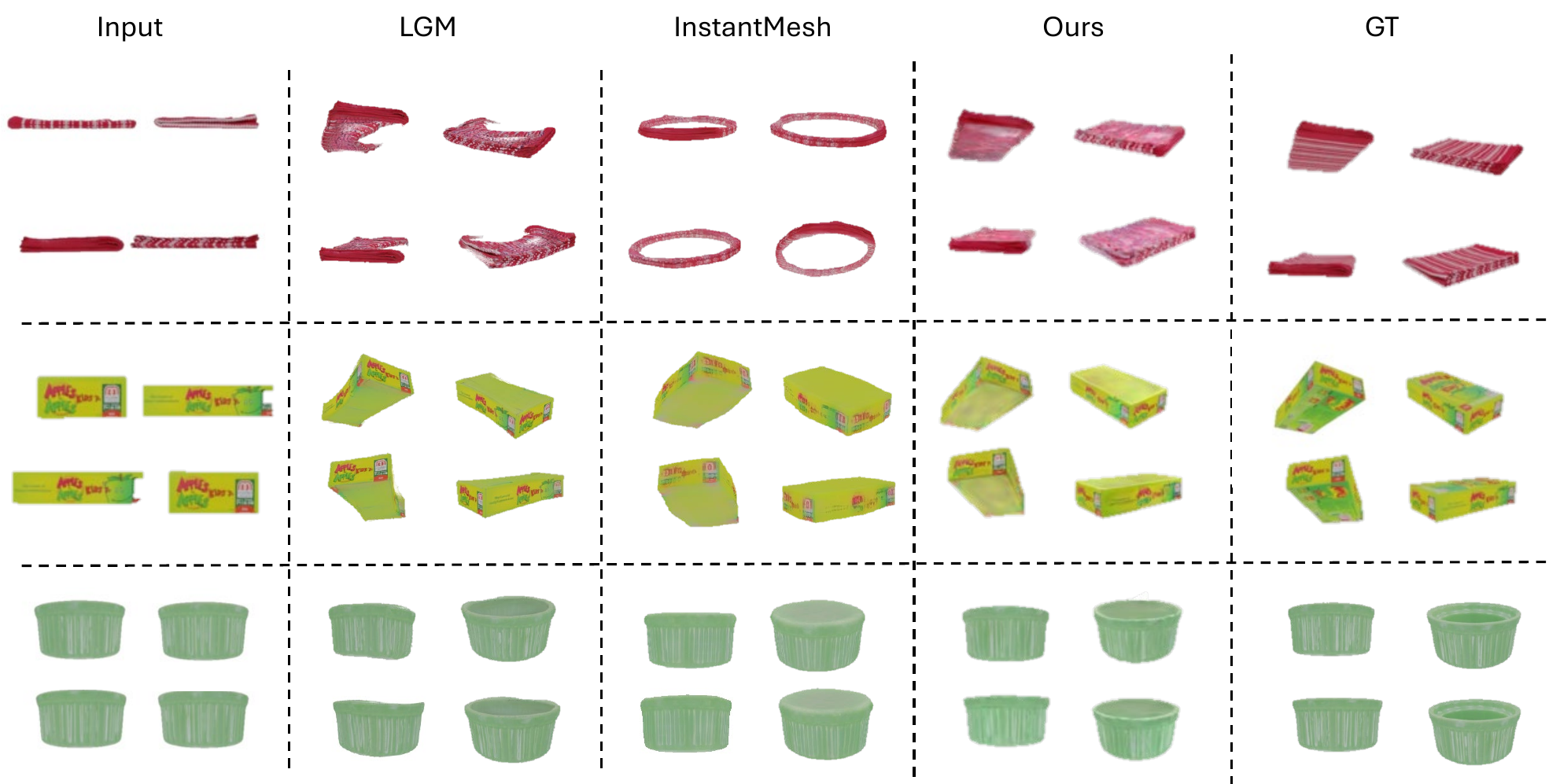}
    \caption{Quality for rendered novel views on GSO dataset for inputting 4 views with resolution 256 LGM large model.}
    \label{fig: fixed_vis_256}
\end{figure*}
As shown in \cref{fig: fixed_vis_256}, when given limited number of input, neither LGM nor InstantMesh gives the meanful geomery.

\begin{figure*}
    \centering
    \includegraphics[width=1.0\textwidth]{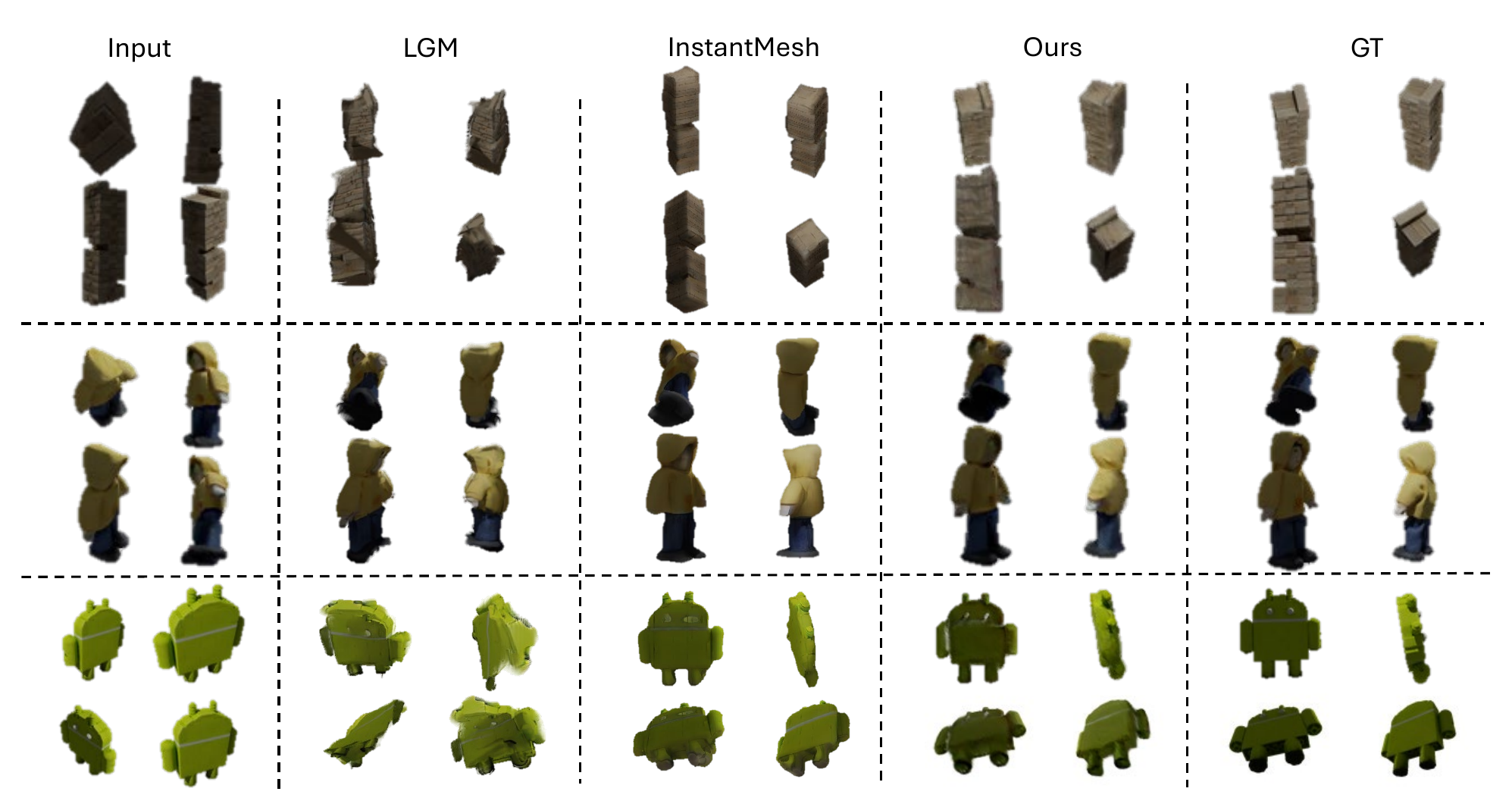}
    \caption{Quality for rendered novel views on GSO dataset for inputting 4 views with random camera poses.}
    \label{fig: random_vis}
\end{figure*}
\cref{fig: random_vis} presents the quantitative results of novel views rendered by recent models trained on 4 views. When provided with 4 random views as input, LGM \citep{LGM} demonstrates a loss of geometry and encounters `ghosting' problems stemming from its training on fixed views. In contrast, our approach produces a cohesive 3D Gaussian set that effectively captures object geometries.

The figures illustrate that LGM encounters the issue of `ghosting'; for instance, there are multiple handles visible for the mushroom teapot. InstantMesh loses some details due to its utilization of a discrete triplane to represent continuous 3D space. 

\begin{figure*}
    \centering
    \includegraphics[width=1.0\textwidth]{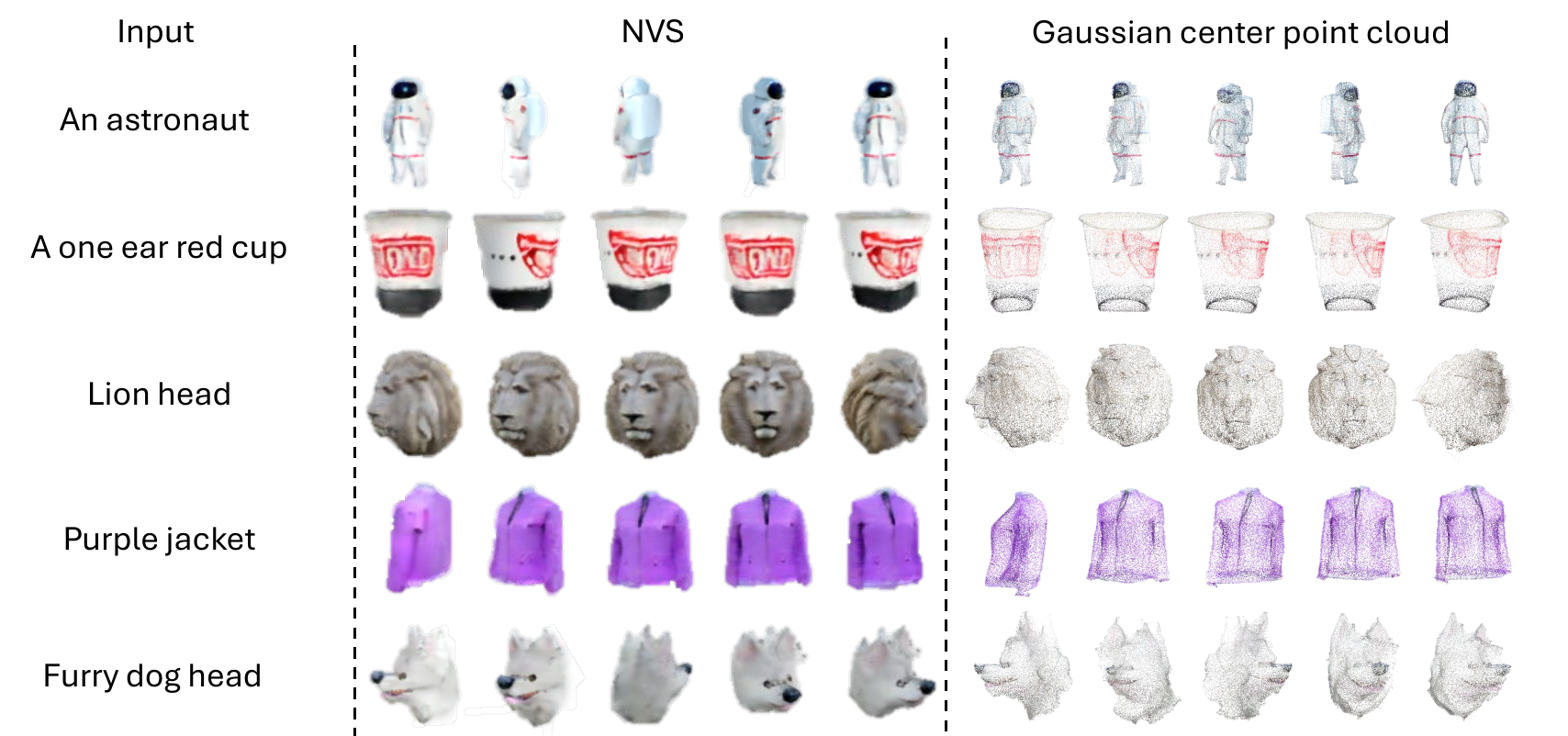}
    \caption{Quality for rendered novel views on inputting text and using MVDream to generate 4 views.}
    \label{fig: from_text}
\end{figure*}

\cref{fig: from_text} shows the result of text-to-3D task. We have incorporated text-to-3D capabilities into our model. To assess quality, we employ MVDream \citep{mvdream} to create a single image from a text prompt. Subsequently, a diffusion model is utilized to generate multi-view images, which are then processed by our model to obtain a 3D representation.

\begin{figure}
    \centering
    \includegraphics[width=0.99\textwidth]{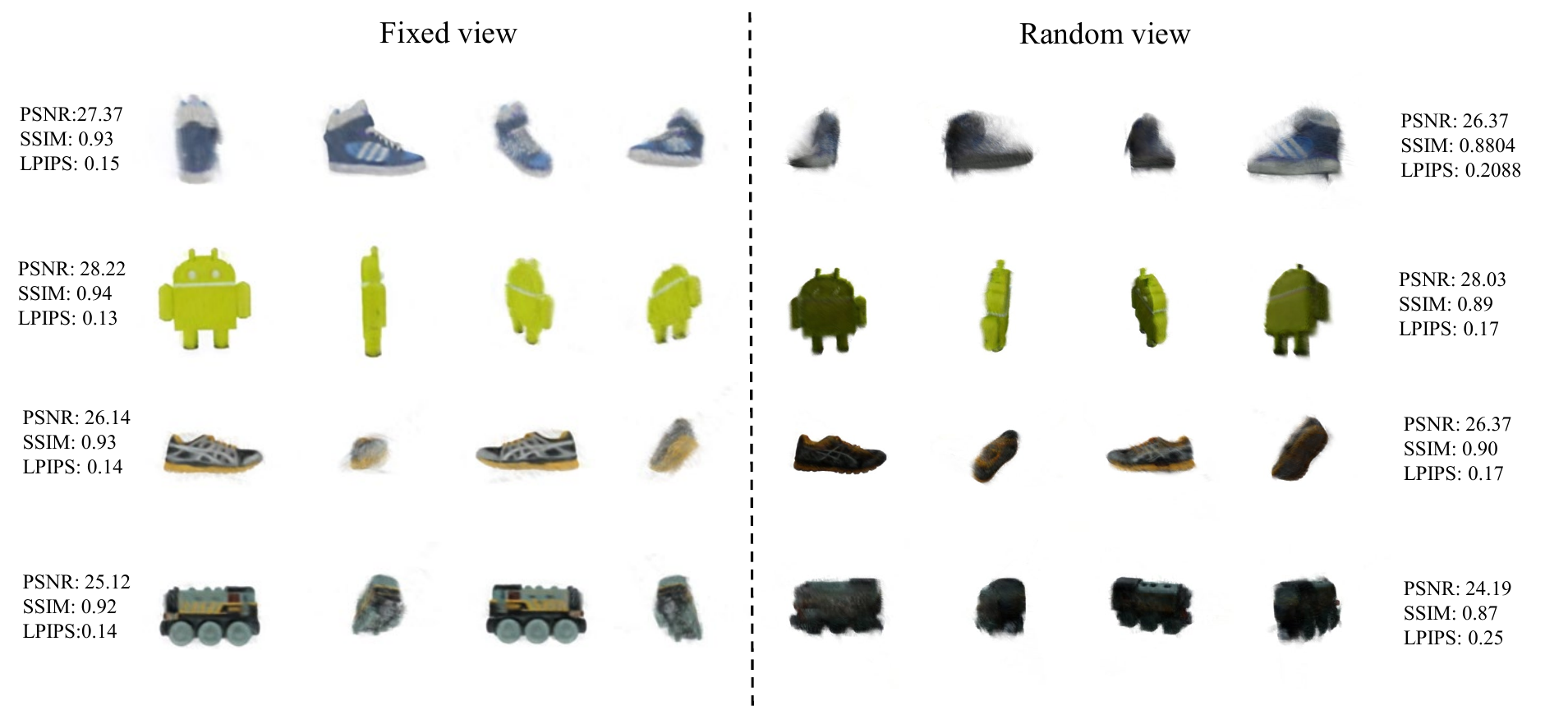}
    \caption{Visualization for Splatter Image with fixed view input and random view input.}
    \label{fig: fix_vs_rand}
\end{figure}

The setting of random input view is obvious a more challenging task than the setting of fixed input view, thus our method also inevitably suffers from a performance drop but still perform better than other state-of-the-art methods. As for Splatter Image \citep{SplatterImage}, it also meets a significant performance drop when random input views are used as its SSIM $\uparrow$ decreased from 0.9151 to 0.8932 and LPIPS $\downarrow$ increased from 0.1517 to 0.2575 despite its PSNR $\uparrow$ has a slight increase. We visualize the results of the two settings to show the difference in \cref{fig: fix_vs_rand}.

\begin{figure*}
    \centering
    \includegraphics[width=1.0\textwidth]{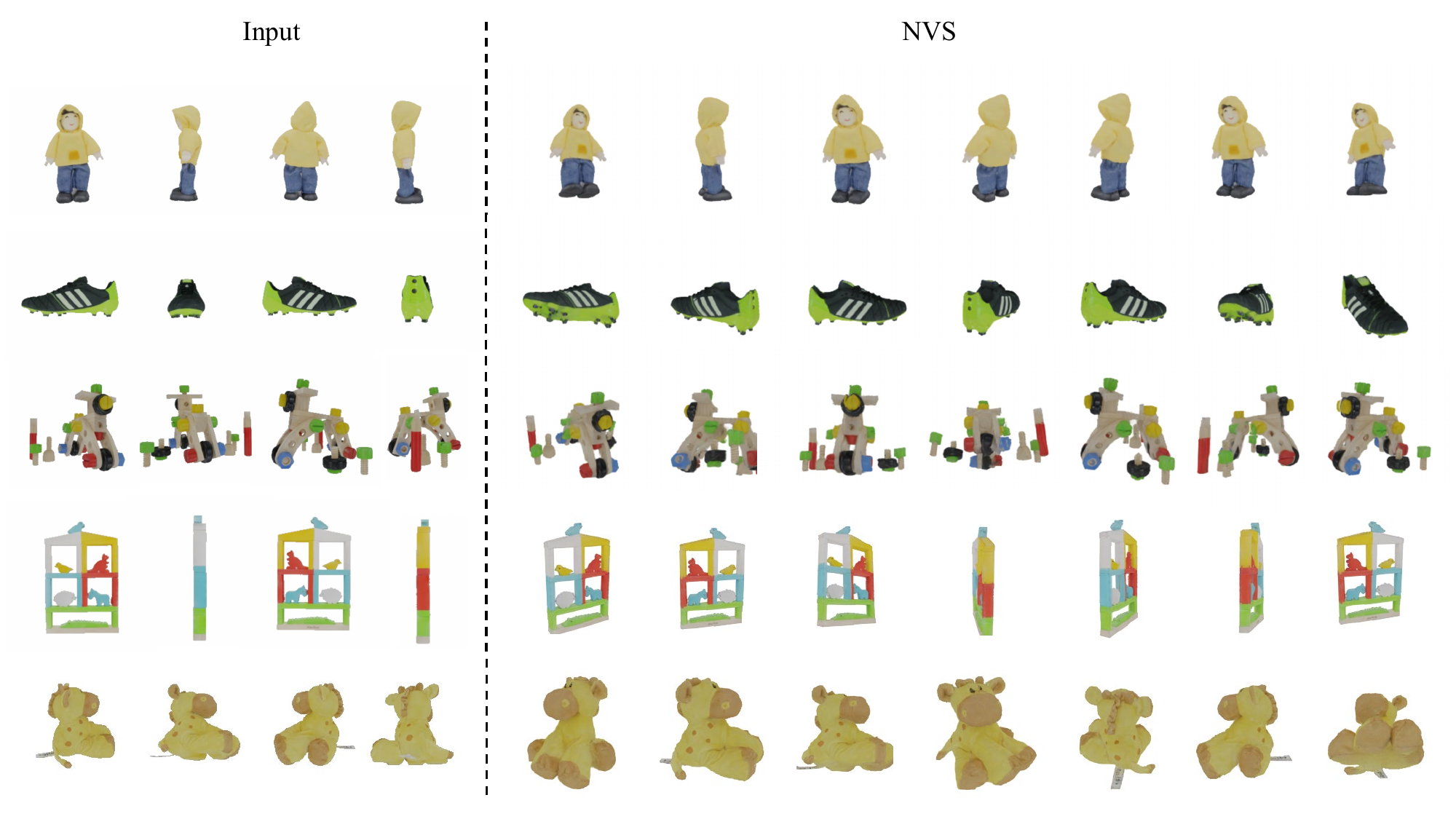}
    \caption{Visualization for our method with resolution 512}
    \label{fig: 512_res}
\end{figure*}

We provide the visualization result with resolution 512 in \cref{fig: 512_res}.


\subsubsection{Single image reconstruction} \label{app: single_view}

\begin{figure*}
    \centering
    \includegraphics[width=1.0\textwidth]{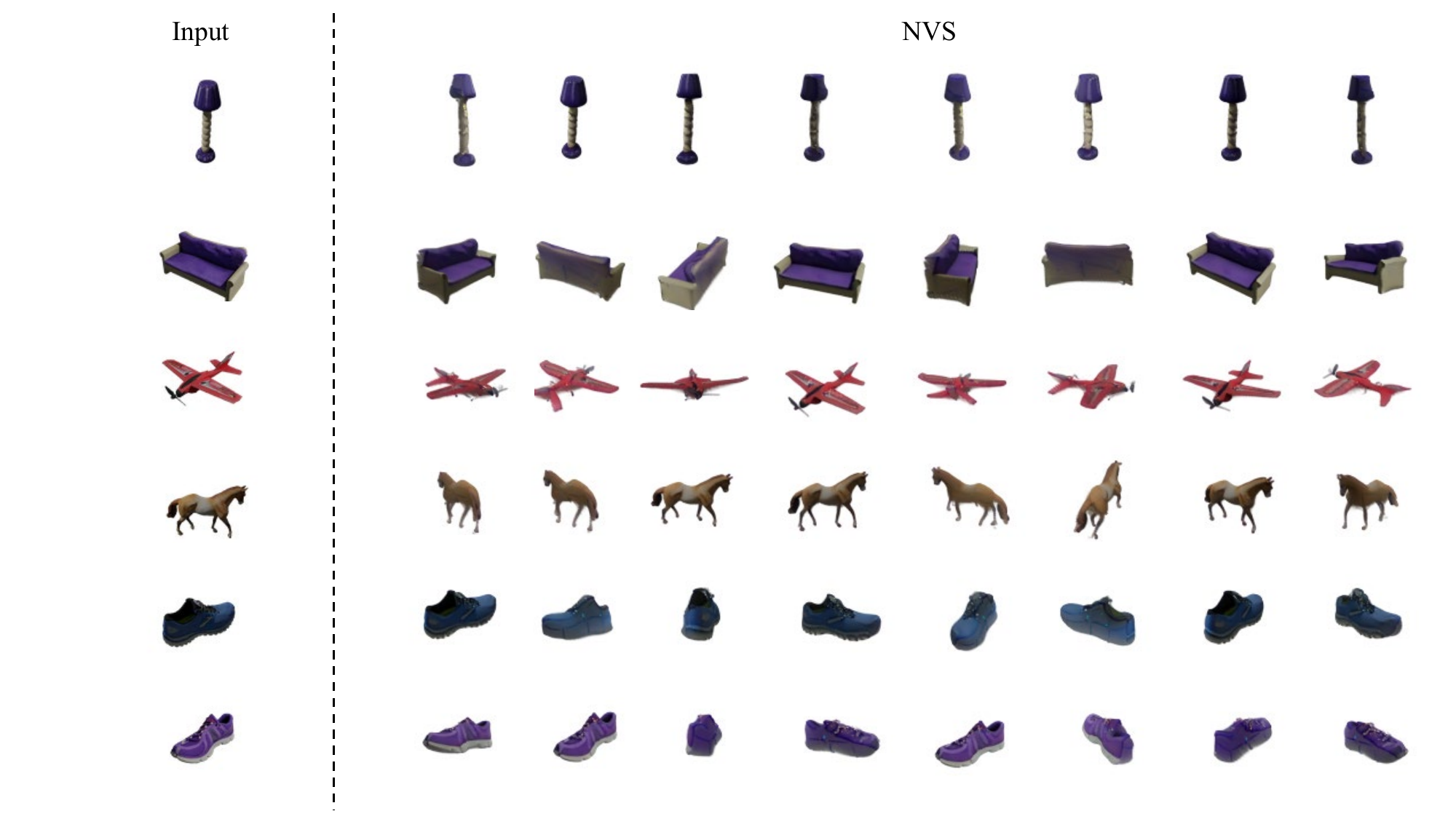}
    \caption{Single view 360 rendering visualization on GSO dataset}
    \label{fig: single_view}
\end{figure*}

\begin{table*}[htp]
  \caption{Quantitative results trained on Objaverse LVIS and tested on GSO. 3D sup. means need 3D supervision.}
  \label{tab: triplane_results}
  \centering
  \scriptsize
  \begin{tabular}{lllllll}
    \toprule
       Method & PSNR $\uparrow$     & SSIM $\uparrow$ & LPIPS $\downarrow$ & 3D sup. & Inference time & Rendering time\\
    \midrule
    Triplane-Gaussian \citep{triplane-gs} & 18.61 & 0.853 & 0.159 & \Checkmark & 1.906 & \textbf{0.0025} \\
    TripoSR \citep{TripoSR} & 20.00 & 0.872 & 0.149 & \XSolidBrush & 3.291 & 22.7312 \\
    \midrule
    Ours & \textbf{23.45} & \textbf{0.897} & \textbf{0.093} &\XSolidBrush & \textbf{0.476}\ & \textbf{0.0025}\\
    \bottomrule
  \end{tabular}
\end{table*}
There are common points between our model and TriplaneGaussian and Instant3D that we all use a unitary representation and use Transformer to regress. For Instant3D, it transformers image to Nerf, making longer rendering time. For Triplane Gaussian, which is a single view reconstruction model with complex and costly triplane representation, representing compresses 3D space, leading to a lack of detailed information in the 3D structure and imposing a rigid grid alignment that limits flexibility \citep{LGM, pointnet}. In the contrast, we use a more efficient way (deformable attention) to decode Gaussians. The comparison between Triplane-Gaussian and our methods is shown in \cref{tab: triplane_results}. Triplane Gaussian requires 3D supervision and takes longer inference time while get worse performance comparing to our model. We test on the given light-weight checkpoint in the github on the single view situation. We also test TripoSR \citep{TripoSR} on the single image reconstruction setting. As shown in \cref{tab: triplane_results}, our model surpass the previous methods on both the performance and the inference speed. We provide the visualization results of our model on single image reconstruction task in \cref{fig: single_view}

\subsubsection{Comparison to masked LGM and Splatter Image} \label{app: mask}
\begin{figure*}
    \centering
    \includegraphics[width=1.0\textwidth]{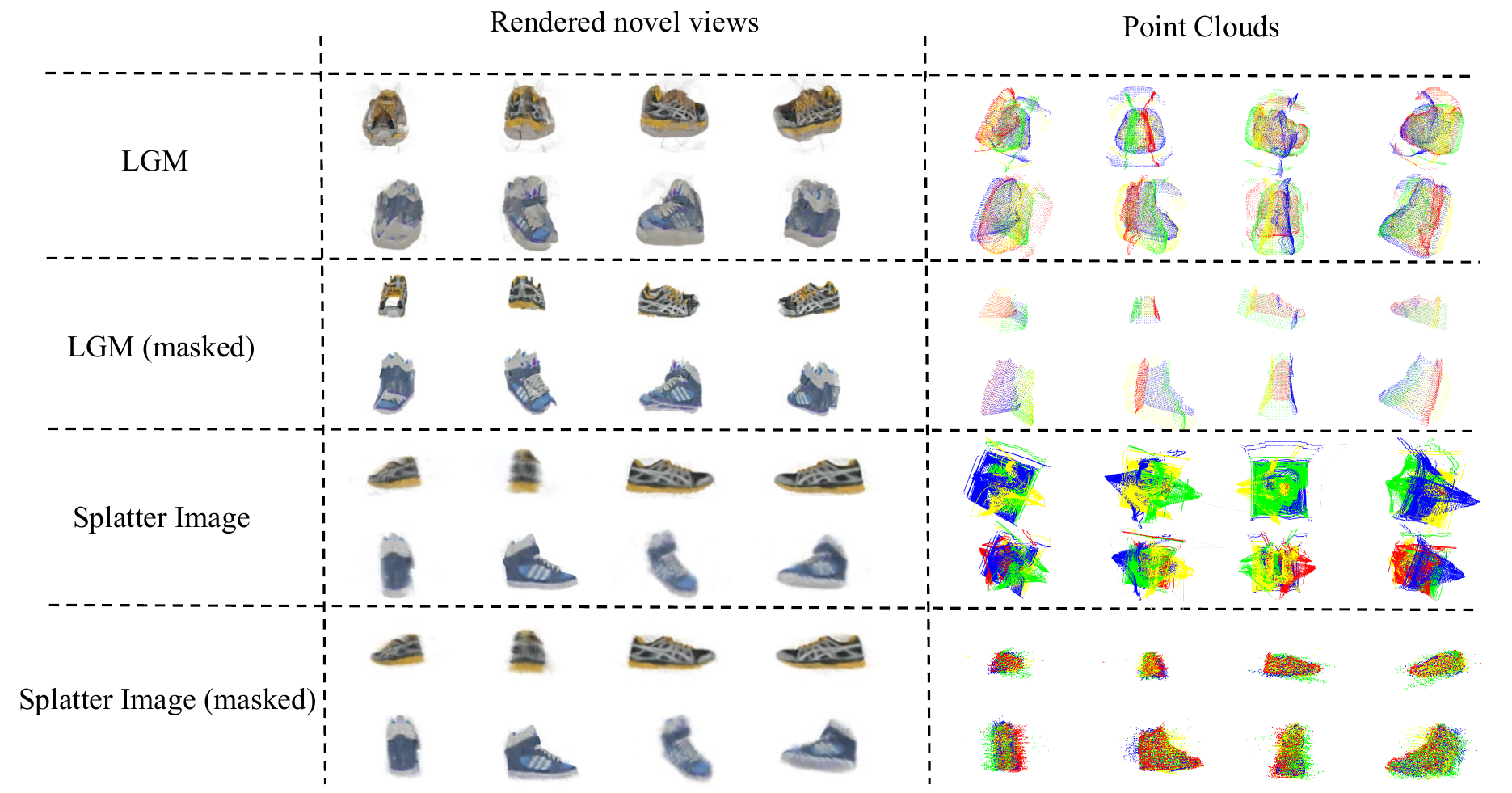}
    \caption{Removing the background use mask for Splatter Image and LGM}
    \label{fig: mask}
\end{figure*}

\begin{table}[htp]
  \caption{Comparison between masked and original pixel aligned methods}
  \label{tab: mask_results}
  \centering
  \footnotesize
  \begin{tabular}{llll}
    \toprule
       Method & PSNR $\uparrow$     & SSIM $\uparrow$ & LPIPS $\downarrow$ \\
    \midrule
    LGM & 17.4810 & 0.7829 & 0.2180 \\
    LGM (masked) & 21.6008 & 0.8608 & 0.1232\\
    Splatter Image & 25.6241 & 0.9151 & 0.1517 \\
    Splatter Image (masked) & 25.0648 & 0.9147 & 0.1684 \\
    \bottomrule
  \end{tabular}
\end{table}

To better explain that the `ghosting' problem is not caused by the background points from previous methods, we provide the results on removing background points of LGM and Splatter Image. LGM uses mask loss to make the most of the pixels contribute to the object itself, even for the background pixels, therefore, removing background use mask makes the results more sparse. It also removing some outliers and thus the rendering results is better as shown in \cref{tab: mask_results}.
Splatter Image keep most of the pixels contribute to its original position, making most of the background points still located on a plane instead of the object. Therefore, removing background use mask does not influence the rendering result much but the rendering quality still reduced a little. 
Moreover, the `ghosting' is not caused by the background points but the mis-alignment of 3D Gaussians from different views, removing the background use mask does not help solving the problem. We show the visualization in \cref{fig: mask}

\subsubsection{Other number of view results} \label{app: other_num_views}

We present the results of training with varying numbers of views (2, 6, 8) and evaluate the corresponding results with the same number of views in \cref{tab: random_view_results}.
\begin{table*}[htp]
  \scriptsize
  \caption{Quantitative results of novel view synthesis training using 2, 6, and 8 input views, tested on the GSO dataset across 2, 6, and 8 views.}
  \label{tab: random_view_results}
  \centering
  \begin{tabular}{llllllllll}
    
    \multicolumn{1}{c}{Method} & \multicolumn{3}{c}{2 views} & \multicolumn{3}{c}{6 views} & \multicolumn{3}{c}{8 views}          \\
         & PSNR $\uparrow$    & SSIM $\uparrow$ & LPIPS $\downarrow$  & PSNR $\uparrow$     & SSIM $\uparrow$ & LPIPS $\downarrow$   & PSNR $\uparrow$     & SSIM $\uparrow$ & LPIPS $\downarrow$ \\
    \midrule
    Splatter Image & 22.6390 & 0.8889 & 0.1569 & 26.1225 & 0.9178 & 0.1620 & 26.4588 & 0.9166 & 0.1714\\
    \midrule
    Our Model & \textbf{23.8384} & \textbf{0.8995} & \textbf{0.1254} & \textbf{28.1035} & \textbf{0.9489} & \textbf{0.0559} & \textbf{28.8262} & \textbf{0.9537} & \textbf{0.0492} \\
    \bottomrule
  \end{tabular}
\end{table*}

\begin{figure}
    \centering
    \includegraphics[width=0.55\textwidth]{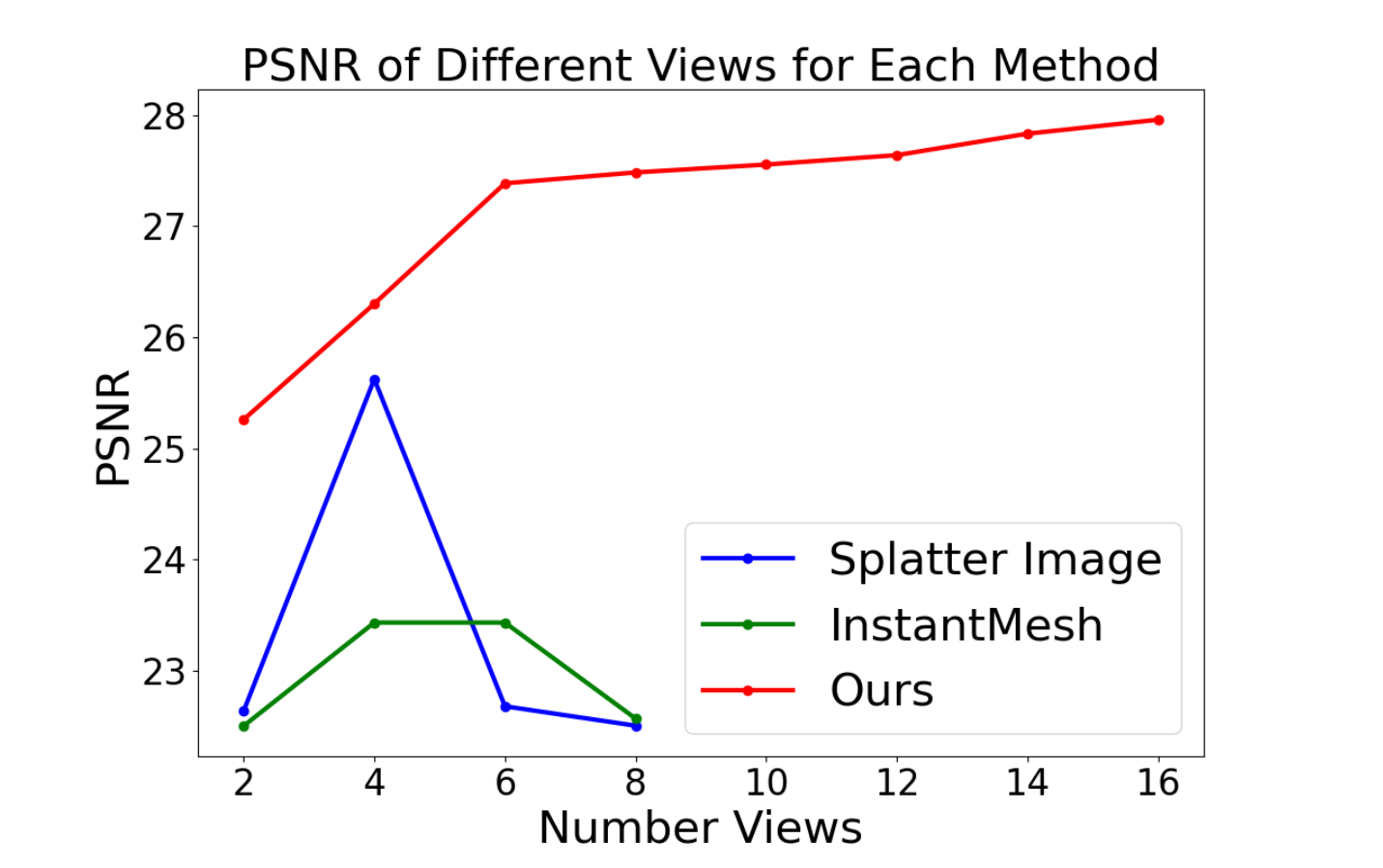}
    \caption{Visualization for Splatter Image with fixed view input and random view input.}
    \label{fig: more_view}
\end{figure}

Our model is positioned on the `sparse view' setting, which indicates the number of views less then 10, so we only reports the performance of views from 2 to 8 in the main paper. With the increase of input views, information from similar views becomes redundant, so the gain for our model has become plateaued while other methods suffer from performance drop as they cannot handle too many input views due to the view inconsistent problem. As we keep increasing the number of input views larger than 8, our method can still benefit from more input views (as shown in \cref{fig: more_view}) while others meet the CUDA-out-of-memory problem.

\subsubsection{Comparison to MVSplat and pixelSplat} \label{app: mvsplat}

We present a comparative analysis involving MVSplat \citep{mvsplat} and pixelSplat \citep{pixelsplat} on the GSO dataset by training it on Objaverse. Similar to LGM \citep{LGM}, both aforementioned methods follow a workflow that regress Gaussians from each views within the respective camera spaces and subsequently merge them in the global world space. Despite pixelSplat's integration of cross-view-aware features through an epipolar Transformer, accurately forecasting a dependable probabilistic depth distribution based solely on image features remains a formidable task \citep{mvsplat}. This limitation often translates to pixelSplat's geometry reconstruction exhibiting comparatively lower quality and plagued by noticeable noisy artifacts \citep{mvsplat}.
Upon examination, we observed that even after isolating points within a visual cone and eliminating background Gaussians, the geometry fails to convey meaningful information, yielding unsatisfactory results. 

In contrast, MVSplat adopts a design that incorporates a cost volume storing cross-view feature similarities for all possible depth candidates. These similarities offer crucial geometric cues for 3D surface localization, leading to more substantial depth predictions. However, akin to Splatter Image, which assigns each pixel a Gaussian and thereby generates a planar representation rather than the object itself, MVSplat's approach may obscure object details due to occlusion by background Gaussians from other viewpoints, resulting in suboptimal outcomes.

\begin{figure*}
    \centering
    \includegraphics[width=1.0\textwidth]{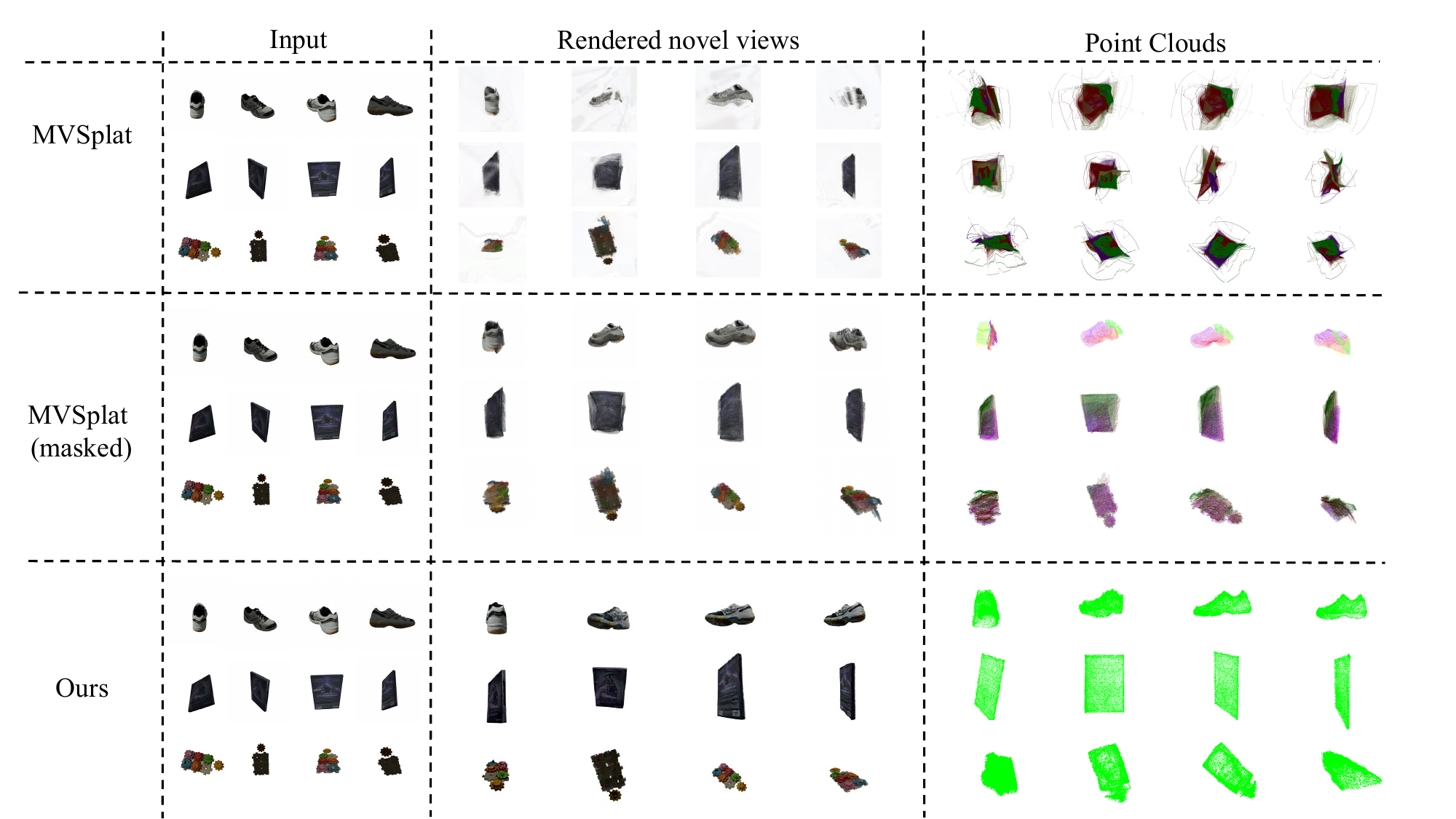}
    \caption{Visualization for MVSplat and our method}
    \label{fig: mvsplat}
\end{figure*}

To address this issue, we selectively mask the positioning of Gaussians on background pixels, focusing solely on rendering Gaussians contributing to the object itself. This adjustment reveals significant `ghosting' problems, as illustrated in \cref{fig: mvsplat}. In the figure, we present the centers of Gaussians generated from different views in different color and the novel views are rendered from the Gaussians from all views. Furthermore, the elaborate incorporation of cross-view attention mechanisms and cost volumes in MVSplat leads to extended inference times and heightened memory requirements as shown in \cref{tab: mvsplat}.

\begin{table*}[htp]
  \caption{Comparison with MVSplat on the GSO dataset in the 4-view input setting.}
  \label{tab: mvsplat}
  \centering
  \begin{tabular}{lllllll}
    \toprule
       Method & PSNR $\uparrow$     & SSIM $\uparrow$ & LPIPS $\downarrow$ & Inference time & Rendering time\\
    \midrule
    MVSplat & 23.06 & 0.90 & 0.13 & 0.112 & 0.0090 \\
    MVSplat (masked) & 24.10 & 0.91 & 0.12 & 0.112 & 0.0045 \\
    \midrule
    Ours & \textbf{26.30} & \textbf{0.93} & \textbf{0.08}  & 0.694 & \textbf{0.0019}\\
    \bottomrule
  \end{tabular}
\end{table*}

\subsection{Ablation study} \label{app: ablation}

\paragraph{Number of views for the initialization}
We add the ablation study on the number of images used during the the training of initialization. The results shown is that the number of images used does not influence the final result. The reason that we choose the number of views being 2 is that we want to support any number of input views. For example, if we choose the number of views being 8, we should at least provide 8 views so that the model can not support the number of views smaller than 8. And we tried to change the input views but the number of input views keeping 2 unchanged, the variance of PSNR for 10 different experiments is within 0.185.

\begin{table}[htp]
  \caption{Ablation study results of different view and different number of views for the initialization (with 4 views in the refinement stage)}
  \label{tab: ablation_n_view}
  \centering
  \footnotesize
  \begin{tabular}{llll}
    \toprule
       Number of views & PSNR $\uparrow$     & SSIM $\uparrow$ & LPIPS $\downarrow$ \\
    \midrule
    1 & 30.2312 & 0.9608 & 0.0413 \\
    2 & 30.4245 & 0.9614 & 0.0422 \\
    3 & 30.3442 & 0.9618 & 0.0419 \\
    4 & 30.4521 & 0.9620 & 0.0412 \\

    \bottomrule
  \end{tabular}
\end{table}
\paragraph{Convergence for different regression target} \label{app: ablation_study_convergence}
Upon investigation, we observe that prior techniques frequently predict depth rather than the centers of Gaussians. In our exploration, we conduct experiments focusing on regressing the centers of 3D Gaussians while keeping other aspects constant. Through this analysis, we discover that regressing the positions of 3D Gaussians can introduce convergence obstacles. Table \cref{tab: convergence} illustrates the outcomes of these experiments on the Objaverse validation dataset after 100K steps.

\begin{table*}[htp]
  \small
  \caption{Ablation study on parameter selection.}
  \label{tab: convergence}
  \centering
  \begin{tabular}{llll}
    \toprule
    Regression target & PSNR $\uparrow$     & SSIM $\uparrow$ & LPIPS $\downarrow$ \\
    \midrule
    Depth & 24.3792 & 0.9012 & 0.1014   \\
    3D Gaussian centers (random initialize in visual cone) & 19.2551 & 0.8343 & 0.1876  \\
    \midrule
    With initialization & \textbf{25.5338} & \textbf{0.9126} & \textbf{0.0833} \\
    \bottomrule
  \end{tabular}
\end{table*}

\paragraph{More ablation studies} \label{app: ablation_study}
Here we gives more ablation study mainly for hyperparameter selection. Due to computational costs, ablation models are trained at 100k iteration and test on Objaverse validation dataset. 
\paragraph{Hyperparameter selection}


\begin{table*}[htp]
  \small
  \caption{Ablation study on parameter selection.}
  \label{tab: ablation_study_param}
  \centering
  \begin{tabular}{llll}
    \toprule
    Method & PSNR $\uparrow$     & SSIM $\uparrow$ & LPIPS $\downarrow$ \\
    \midrule
    2 decoder layers & 24.5229 & 0.9195 & 0.1021   \\
    6 decoder layers & \textbf{26.2442} & \textbf{0.9352} & \textbf{0.0778}  \\
    Freeze encoder & 25.3211 & 0.9264 & 0.1003 \\
    \midrule
    Default model & 26.2313 & 0.9351 & 0.0788   \\
    \bottomrule
  \end{tabular}
\end{table*}

In \cref{tab: ablation_study_param}, we opted for 4 decoder layers over 6, as the latter offers marginal improvement but demands significantly more computational resources. Our findings indicate that fine-tuning yields the better results.

\end{document}